\documentclass[twoside,11pt]{article}

\usepackage[accepted]{melba}
\usepackage{comment}
\usepackage{dirtytalk}
\usepackage{tabularx}
\usepackage[absolute]{textpos}
\usepackage{dsfont}
\usepackage{multirow}
\usepackage{multicol}
\usepackage{amsmath}
\usepackage{mwe} 

\usepackage{amsmath,amsfonts}

\melbaid{2024:005}  
\doi{https://doi.org/10.59275/j.melba.2024-d434}
\melbaauthors{Shuaibu and Simpson}  
\volume{2}
\firstpageno{686}  
\melbayear{2024}  
\datesubmitted{10/2023}  
\datepublished{03/2024}  

\melbaspecialissue{Special Issue on Image Registration}
\melbaspecialissueeditors{Veronika Zimmer, Alessa Hering, Mattias Heinrich}

\ShortHeadings{HyperPredict: Estimating Hyperparameter Effects in DIR}{Shuaibu and Simpson}

\title{HyperPredict: Estimating Hyperparameter Effects for Instance-Specific Regularization in Deformable Image Registration.}

\author{\firstname Aisha L. \surname Shuaibu \orcid{https://orcid.org/0000-0002-6061-2231}\email a.shuaibu@sussex.ac.uk \\  
	\addr University of Sussex, Predictive Analytics Lab (PAL)
	\AND
	\name Ivor J. A. \surname Simpson\orcid{0000-1111-2222-3333} \email i.simpson@sussex.ac.uk \\
	\addr University of Sussex, Predictive Analytics Lab (PAL)
}

\begin{document}

\maketitle

\begin{abstract}
Methods for medical image registration infer geometric transformations that align pairs, or groups, of images by maximising an image similarity metric. This problem is ill-posed as several solutions may have equivalent likelihoods, also optimising purely for image similarity can yield implausible deformable transformations. For these reasons regularization terms are essential to obtain meaningful registration results. However, this requires the introduction of at least one hyperparameter, often termed $\lambda$, that serves as a trade-off between loss terms. In some approaches and situations, the quality of the estimated transformation greatly depends on hyperparameter choice, and different choices may be required depending on the characteristics of the data. Analyzing the effect of these hyperparameters requires labelled data, which is not commonly available at test-time. In this paper, we propose a novel method for evaluating the influence of hyperparameters and subsequently selecting an optimal value for given pair of images. Our approach, which we call HyperPredict, implements a Multi-Layer Perceptron that learns the effect of selecting particular hyperparameters for registering an image pair by predicting the resulting segmentation overlap and measures of deformation smoothness. This approach enables us to select optimal hyperparameters at test time without requiring labelled data, removing the need for a one-size-fits-all cross-validation approach. Furthermore, the criteria used to define optimal hyperparameter is flexible post-training, allowing us to efficiently choose specific properties (e.g. overlap of specific anatomical regions of interest, smoothness/plausibility of the final displacement field). 
We evaluate our proposed method on the OASIS brain MR standard benchmark dataset using a recent deep learning approach (cLapIRN) and an algorithmic method (Niftyreg). Our results demonstrate good performance in predicting the effects of regularization hyperparameters and highlight the benefits of our image-pair specific approach to hyperparameter selection.

\end{abstract}


\section{Introduction}
\label{sec:Introduction}

Deformable image registration has been an active field of research for decades, as it is a fundamental process utilized in various medical imaging studies. The process involves aligning images of different modalities and time points for comparison and measurement of changes over time. It is very common for medical images to vary in their spatial resolution and orientation; as a result, non-linear image registration plays a vital role in various clinical applications such as image-guided treatment delivery, pre-operative/post-operative assessment comparison, disease monitoring, disease diagnosis, and population analysis \citep{meng2022enhancing}.


In medical images, the quality of registration largely depends on two conditions: (a) accurate alignment of anatomical structures, (b) a smooth and plausible displacement field. Achieving a balance between these factors is accomplished by utilizing hyperparameters that act as a trade-off between both objectives. Hence, selecting optimal hyperparameter values is a critical aspect when evaluating image registration methods.

The most quantitative method of selecting these hyperparameters is by performing grid or random search on the validation data using a discrete set of hyperparameter values. In this approach, additional data containing anatomical annotations is employed. By computing segmentation overlap of anatomical regions between the ground truth labels and the inferred registration, an \say{optimal} value is selected based on the hyperparameter that maximizes a criteria \emph{across the entire population}. We argue that this approach may lead to sub-optimal choice as data-specific optimal hyperparameter differ significantly depending on various factors such as, the morphological similarity between the images and the anatomical structure of interest, among other factors. Furthermore, researchers may be inclined to adopt values from existing literature that may not be suitable for their specific dataset or registration task, equally leading to sub-optimal results.


%



In this paper, we present a novel method, HyperPredict, an efficient tool for instance-specific optimal hyperparameter selection in registration tasks. Our proposed approach enables greater flexibility at test time and can be leveraged in scenarios where labelled scans are scarce. In summary our contributions are as follows:

\begin{itemize}
    \item We propose a novel and efficient method of \emph{learning} the effect of registration hyperparameters. This is achieved by learning the parameters of a network that maps a set of hyperparameters and image pair to desired evaluation metrics (derived from segmentation overlap and smoothness of deformation field). 
     
    \item Our method is flexible and allows the efficient choice of optimal parameters based on a defined criteria.
    
    \item We test our proposed method using two different registration algorithms (cLapIRN and Niftyreg), our code is publicly available at ~\url{https://github.com/aisha-lawal/hyperpredict}.


\end{itemize}

The structure of the paper is as follows. Section 2 introduces deformable image registration, it's mathematical formulation and hyperparameters in medical image registration tasks, Section 3 describes related work. In Section 4 we present our method. Sections 5 and 6 describe experimental results and discuss insights. Finally, we present limitations of our method and conclude in Sections 7 and 8 respectively.

\section{Background}
\label{sec:Background}
\noindent\textbf{Deformable Image Registration: } Conventional registration algorithms \citep{ashburner2007fast, beg2005computing, avants2008symmetric, niftyreg} solve the optimization problem for each volume pair by iteratively improving estimates for the desired transformation such that the loss function is minimized. However, solving a pairwise optimization problem is computationally expensive and can be very slow in practical medical applications. Over the years, deep learning image registration methods (DLIR) \citep{LapIRN, cLapIRN, voxelmorph} have been introduced to circumvent this by optimizing the parameters of a network, the general process aims to establish a dense, non-linear correspondence between pairs of images, such as magnetic resonance (MR) images or computerised tomography (CT) scans, by learning the optimal spatial transformation between the image pair that enhances similarity. The process of registering two images can be formulated as an optimization problem generally expressed as: 


\begin{equation}
\label{eqn:general_formualation}
{\phi^*} =  \underset{\phi}{\arg\min} \mathcal{L}_{\text{sim}}(f ,m\circ\phi)
\end{equation}
Equation \ref{eqn:general_formualation} seeks to minimize a loss that consists of two parameters, \emph{f} and \emph{$m\circ \phi$}. Where \emph{m} and \emph{f} denote the moving and fixed image respectively, \emph{$\phi$} represents the deformation field, \emph{$\phi^*$} is the optimal  registration field that maps the pixels/voxels from \emph{m} to \emph{f}.

Intuitively, deformable image registration poses a significant challenge due to its inherently ill-posed nature. Reliance solely on surrogate measures such as image similarity can be insufficient as they do not have the ability to differentiate between accurate and inaccurate registrations \citep{rohlfing2011image}. For example, given an image pair, \emph{m} and \emph{f}, registration algorithms seek to find the transformation that deforms the moving image, mapping the coordinates of \emph{m} to \emph{f}; however, ground-truth deformation fields do not exist to serve as a reference point, and without any restrictions placed on the transformation properties, the cost function becomes poorly conditioned. Thus, to ensure tractability, registration algorithms employ some regularization that imposes a constraint on the estimated deformation field - reducing the set of possible solutions. The quality of the registration largely depends on the regularization weight, \emph{$\lambda$}, that serves as a trade-off between the quality of the registration and how smooth the deformation field is. Thus, the cost function can be reformulated as follows:

\begin{equation}
\label{eqn:conditioned_formualation}
{\phi^*} = \underset{\phi}{\arg\min} \mathcal{L}_{\text{sim}}(f ,m \circ \phi) + {\lambda} \mathcal{L}_{\text{reg}}(\phi)
\end{equation}

Description of the notation is similar to Equation \ref{eqn:general_formualation}, with $\lambda$ denoting the regularization weight,  $\mathcal{L}_{\text{sim}}$ and $\mathcal{L}_{\text{reg}}$ represent the dissimilarity and imposed regularization function respectively. The aim is to optimize both the registration quality, $\mathcal{L}_{\text{sim}}$, while having a smooth deformation field, $\mathcal{L}_{\text{reg}}$. The choice of loss function for $\mathcal{L}_{\text{sim}}$ depends on various factors including the intensity distribution and contrast of the image. Commonly used functions are mean squared error, mutual information \citep{mutualinformation}, and normalized cross correlation \citep{NCC}, which capture different aspects of similarity. For $\mathcal{L}_{\text{reg}}$, a diverse range of regularisers can be employed to enforce spatially smooth deformations, examples include,  linear elasticity, bending energy, and learned priors. \\

\noindent\textbf{Evaluation Metrics:} Quantifying the accuracy of non-rigid image registration is inherently difficult. As such, no gold standard for evaluating deformable registration exists. However, with \emph{additional labelled data}, we can make an effort to quantify the quality of the registration with some degree of confidence. A common approach as a means of evaluation is to compute the segmentation overlap of different anatomical regions between \emph{m} and \emph{f}. The Dice score \citep{Dice1945measures} is one way to do this. While the Dice score between anatomical structures is a reliable proxy for evaluation, a high Dice score does not imply a biologically plausible registration.  This is because a deformation field with overlapping voxels can still yield a high Dice score despite potentially unrealistic deformations \citep{rohlfing2011image}. Therefore, in addition to the Dice score, assessing the deformation's diffeomorphic property is equally important in order to preserve the topology of the features in the transformed image. Hence, the determinant of the Jacobian $|J_{\phi}|$ serves as a measure of smoothness of the deformation field.

The experiments presented in this paper employ the Dice score and number of folded voxels (derived from $|J_{\phi}|$) as the desired evaluation metrics. However, it is worth emphasizing that our methodology is not limited to these specific choices. Our approach is flexible and can be easily extended to accommodate a wide range of hyperparameters and evaluation metrics. Depending on specific application or research objective, our method can be adopted to fit the desired metrics, such as Hausdorff distance and Target Registration Error (TRE) amongst others.

\noindent\textbf{Hyperparameter Selection: }Hyperparameter optimization algorithms tackle the challenge of jointly optimizing both model hyperparameters and model weights through a validation and training objective respectively. The simplest approach involves treating model training as a \say{block-box} function and employing methods such as grid search, random search, or sequential search \citep{bergstra2012random}. Other methods include manual fine-tunning and Bayesian optimization \citep{bergstra2011algorithms, turner2021bayesian, snoek2012practical, mockus1998application}. Although effective, these methods can be inefficient as they require repetitive optimization procedures for each hyperparameter value. We describe existing methods in Section \ref{sec:Related Work}.

\noindent\textbf{Hyperparameters in Registration: }In non-rigid image registration, the number of hyperparameters to be optimized depends on the specific registration algorithm and objective. For example, cLapIRN, \citep{cLapIRN} regulate the smoothness of the deformation field using a \emph{single} registration hyperparameter, denoted as $\lambda$. On the other hand, algorithms such as ConvexAdam, \citep{siebert2021fast} and Niftyreg, \citep{niftyreg} employ a \emph{set of} hyperparameters to optimize the registration process. In Niftyreg, spacing for spline interpolation, bending energy, and linear elasticity are some of the regularization options to govern the diffeomorphic properties of the deformation field. In such scenarios, especially when dealing with algorithms with high computational complexity, employing the above methods to tune multiple parameters becomes impractical.

Motivated by the challenge above, HyperPredict presents a more efficient method for selecting optimal hyperparameter values. During training, HyperPredict learns the effect of the hyperparameter on the evaluation metrics. By utilizing a registration algorithm, the target values of both metrics (described above) are obtained for backpropagation. At test time, given an input \{\emph{m, f, $\lambda$}\}, and without having true segmentation, the model predicts the metrics associated with the input. Based on a specific criteria (defined in method section), we select optimal parameter, $\lambda^*$, and use that for registration.

\section{Related Work }
\label{sec:Related Work}
As described in Section \ref{sec:Background}, registration methods typically optimize a data term and a weighted regularization term. Hence, to be able to minimize the error of the registered results, it becomes crucial to optimize the independent hyperparameters of the registration model. Despite the advantages that come with DLIR methods, they still face challenges in navigating the trade-off parameter within the objective function. Traditionally, the selection of regularization parameter values relied on a trial and error approach \citep{ashburner2007fast, andersson2007non, rueckert1999nonrigid}. In this, users manually search for parameters that yield satisfactory results for a given dataset. This process often involves fine-tuning multiple parameters, which can be time consuming, particularly in hierarchical registration methods. An example is studies conducted in \citep{ruhaak2017estimation}, their registration method utilized four distinct parameters. To capture the effect of each parameter, they conducted separate experiments, in each, one parameter was varied while the other three remained fixed. This iterative process allowed them to empirically determine an \say{optimal} value for all four parameters. Although ad hoc parameter-tuning may produce satisfactory outcomes, it requires expert domain knowledge to effectively guide the tuning process \citep{joshi2004unbiased, vialard2011diffeomorphic, ma2008bayesian, wang2019data}.


Cross-validation is another standard practice employed for selecting registration hyperparameters \citep{voxelmorph, mok2020fast}. Using grid search, the parameters that yield the best performance on a given dataset is selected for subsequent registrations. As a result, it utilizes a \emph{fixed} level of regularization across the entire set of image pairs, with the assumption that all image pairs require the same degree of regularization. Additionally, this process necessitates availability of manually labelled dataset that accurately represents the testing data of interest. This method of parameter selection requires substantial computational resources and human effort, which may result in sub-optimal parameter choices. 

Employing a hierarchical Bayesian model to infer the regularization parameter is another method demonstrated in studies by \citep{risholm2012selection} and \citep{simpson2012probabilistic}. The general strategy of Bayesian approach leverages a probabilistic model to explore and evaluate the performance of hyperparameters that result in improved registration. This is achieved by characterising the posterior distribution using techniques like Markov Chain Monte Carlo (MCMC) or Variational Inference. \citep{risholm2013bayesian} considered scenarios where the confidence of both observed data and model priors are unknown, aiming to tackle the challenge of finding an objective trade-off between the two terms. They characterised the posterior distribution using Metropolis-Hastings and treated the hyperparameters as latent variables approximately marginalized over. A similar approach is proposed in \citep{wang2021bayesian, zhang2013bayesian}. Given sufficient samples, MCMC yields a good characterization of the posterior, however, the computational demands and complexity of Markov chain is a restricting factor that limits the feasibility of this approach. While Variational Bayes(VB) has less computational burden, it may compromise on the quality of estimates. \citep{simpson2015probabilistic} utilized a spatially adaptive prior to limit unwanted regularization on the estimated transformation. One limitation is that the uncertainty estimates quantify the variability in the displacement field based on the inferred hyperparameters, but they do not account for the uncertainty in the registration caused by variations in the hyperparameters themselves \citep{le2016quantifying}.


Recent advancements in deep learning registration methods such as HyperMorph, \citep{hoopes2021hypermorph} and cLapIRN, \citep{mok2021conditional} propose an approach that eliminates the need for repeatedly training models in search of hyperparameters that boost model performance. They do this by learning the effect of hyperparameters on the deformation field. Specifically, cLapIRN proposes a conditional image registration method that learns the conditional features that are strongly correlated with specific regularization values. On the other hand, HyperMorph leverages a secondary network to generate conditioned weights for the primary network, in order to learn the impact of registration hyperparameters on the deformation field. To quantify the choice of hyperparameter (selected arbitrarily), registration methods require labels. Rather than learning the entire registration process, \citep{niethammer2019metric}, focus on learning a spatially adaptive regularizer within a registration model to preserve the desired level of regularity, however, this is not done on a pair-wise basis. A similar method is adapted in \citep{vialard2014spatially} using a learning based approach.

To summarize the challenges above; depending on the selected method, tuning and selection of optimal hyperparameters involves dealing with one or more of the following (a) computational expense, (b) time consuming nature of the process, (c) requires labelled data at test time which involves human effort and may not readily available, (d) ineffective in selecting optimal hyperparameter (assumes a fixed value throughout the entire dataset). While both amortized inference and algorithmic approaches to registration allow for hyperparameter selection at test time, efficiently selecting an optimal value at test time remains a challenge, which we aim to solve with our proposed HyperPredict.

We emphasize that the goal of HyperPredict is not to function as a registration tool or to directly compare it with existing ones, but rather to serve as a means of aiding registration algorithms in selecting appropriate hyperparameters.

\section{Method }
\label{sec:method}
Current unsupervised learning-based image registration methods that learn the effect of hyperparameters define a network $h_{\psi}$(\emph{m}, \emph{f}, \emph{$\lambda$}) =  $\phi$, where $h_{\psi}$ is parameterized by a convolutional neural network (CNN).  Our presented approach takes advantage of these registration algorithms but uniquely learns a function $g_{\theta}$ that captures the correlation between a hyperparameter value and the evaluation metric for an image pair. Given $e_o$(an encoded representation of \emph{m} and \emph{f}) and sampled hyperparameter value, $\lambda$ (drawn from a log-normal distribution, i.e., $\lambda \sim \mathrm{LogNormal}(\mu, \sigma)$), we parameterize our proposed method as a function, $g_{\theta}(e_{o},\lambda)$ with a Multi-Layer Perceptron (MLP). The proposed method works with any registration algorithm that incorporates a regulariser in its smoothness during optimization. 

\begin{figure}[ht!]
    \label{fig:overview}
    \centering 
    \includegraphics[width = \linewidth]{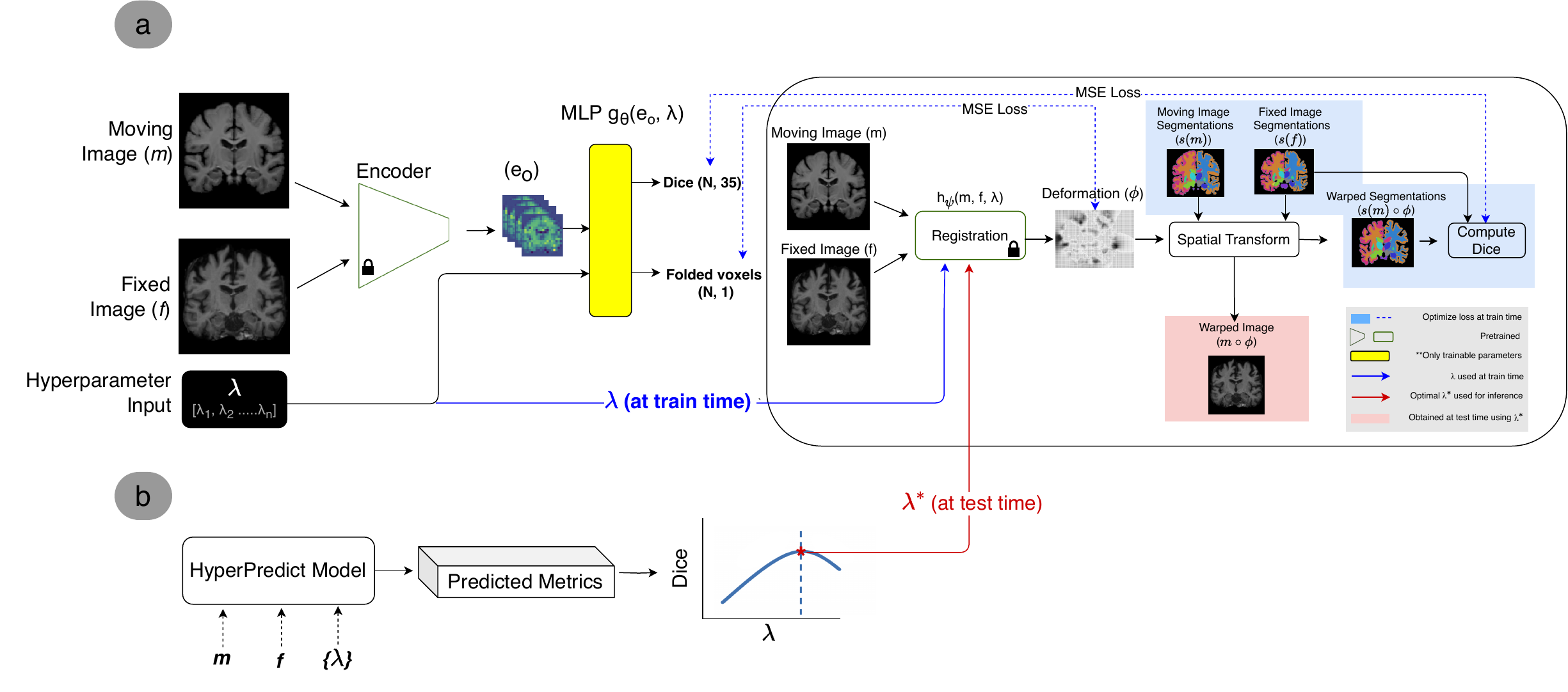}

    \caption{Overview of the method: \textbf{(a)} We learn the parameters $g_{\theta}$ (\textbf{yellow box}) of an MLP that maps the encoded representation ($e_{o}$) of an image pair (\emph{m, f}) and a set of hyperparameters ($\lambda$) to evaluation metrics (Dice and folded voxels). At train time,  we optimize $g_{\theta}$ by computing the target metrics of the input (\textbf{blue boxes and dashed blue lines}) using a registration algorithm. \textbf{(b)} At test time, given a set of input, we predict the effect for each hyperparameter value. We select the optimal hyperparameter, $\lambda^*$, and use that in the registration algorithm to obtain a desired warped image (\textbf{red box}). The green shapes (encoder and registration) are pretrained.} 

\end{figure}

Figure \hyperref[fig:overview]{1} presents an overview of our method. The network architecture during training is comprised of an encoder, a registration algorithm, and the MLP. At train time, we freeze the pretrained weights of the encoder and registration algorithm. Based on the encoded representation and hyperparameter passed as input, the MLP learns to predict the Dice coefficient and number of folded voxels as the evaluation metrics. We use the deformation field, \emph{$\phi$},  from the registration algorithm and the Dice score between the warped moving label, \emph{s(m)}$\circ$ $\phi$, and fixed label, \emph{s(f)}, to enable the computation of the loss. At test time, given unseen images and regularization weight, \{\emph{m, f, $\lambda$}\}, we obtain the predicted metrics by evaluating $g_{\theta}$($e_{o}$, $\lambda$). All notations are illustrated in Figure \hyperref[fig:overview]{1}.

When inferring the optimal weight, $\lambda^*$ (as in Figure \hyperref[fig:overview]{1}(b)), we can impose selection criteria. For instance, we can consider only results that restrict significant discontinuities in the deformation field. This is achieved by constraining the percentage of number of folded voxels (nfv) allowed. Based on this constraint, the optimal weight, $\lambda^*$, that corresponds to the highest Dice score is obtained (across image pairs or selected labels). Mathematically, this can be expressed as;

\begin{equation}
\label{eqn:constraint}
\lambda^* = \underset{\lambda}{\arg\max}(\text{Dice}(\lambda) \mathds{1} [\text{nfv} < \text{N0.5\%}])
\end{equation}
Equation \ref{eqn:constraint} determines the optimal value, ($\lambda^*$) by maximizing the Dice score while ensuring \%nfv is less than 0.5\% of total folded voxels (N). The derived $\lambda^*$ is used for registration. In the following sections, we describe our method in detail.

\subsection{HyperPredict Architecture}
\label{sec:HyperPredict}
This section describes the architecture used in our experiments. Let moving image, \emph{m} and fixed image, \emph{f} represent an image pair, each defined over a 3D spatial domain, $\Omega$ $\subset$ $\mathbb{R}^3$. Since CNNs capture abstract representation of images \citep{szeliski2022computer}, we leverage this to define our encoder. The encoder takes \emph{m} and \emph{f} as input and outputs an encoded representation, \emph{$e_o$}, of the image pair. For simplicity, our experiments utilize two pre-trained registration models as encoders; Symnet, \citep{mok2020fast} and cLapIRN, \citep{cLapIRN}. It is important to note that we need an encoder that could take in a pair of 3D-volumes, the conventional approach of dealing with this is by utilizing a registration model. We performed preliminary experiments using 2D implementation of masked autoencoder \citep{he2022masked} to learn a self-supervised encoding, we found this very slow since they are not designed to work with 3D data. 

The encoder is represented as $\mathbb{R}^{N,C_m,D_m,H_m,W_m} \text{x}   \ \mathbb{R}^{N,C_f,D_f,H_f,W_f} \rightarrow \mathbb{R}^{N,C_{e_o},D_{e_o},H_{e_o},W_{e_o}}$. With \emph{N, C, D, H, W}, indicating the batch size, channels, depth, height and width respectively. The subscripts \emph{m, f, $e_o$}, denote moving, fixed and encoded representation respectively. Using Symnet as an encoder, the resulting representation, \emph{$e_o$} has dimensions of 1x56x10x12x14. We flatten the last three dimensions (D,H,W) and compute the mean across the channels. The result is concatenated with $\lambda$ and passed as input to the MLP. We explored other summary statistics, such as concatenating the mean, max, and min across the channels, comparable results are presented in Appendix C.



Throughout our experiments, we parameterize the function $h_\psi(.,.)$ with two different registration algorithms; cLapIRN, and Niftyreg. Both of which accept hyperparameters as input that condition the smoothness of the registered image. At each iteration, the same set of image pair and hyperparameter is passed to the encoder and registration algorithm simultaneously, the loss is computed between the predicted values from the MLP and target values from the registration.

\subsection{Loss Functions}

In this section, we describe the loss function and evaluation metrics of the model. We optimize the MLP parameters, $g_{\theta}$ using stochastic gradient descent to minimize the loss. Mathematically, the objective function at each iteration is defined as;  

\begin{equation}
\label{eqn:lg}
  \emph{$L$} = \mathcal{L}_{\text{overlap}} + \alpha \mathcal{L}_{\text{nfv}}
\end{equation}
$\mathcal{L}_{\text{overlap}}$ and $\mathcal{L}_{\text{nfv}}$ are the two components of the loss \emph{$L$}, representing the Dice loss and  nfv loss respectively. This is a multitask learning problem, hence we define $\alpha$ as a weighting term, balancing the relative magnitude of both losses, Appendix F shows the relative effect of training with different values of $\alpha$. From the figure, 1.0 proves to be optimal for learning the metrics of both cLapIRN and Niftyreg registration methods. Both components of the loss are evaluated using the mean squared error defined in Equations \ref{eqn:overlap} and \ref{eqn:nfv} below.


Assessing the quality of registration is non-trivial due to competing objectives. Low regularization parameter can allow for close matching of appearance at the cost of anatomically-implausible and highly irregular deformations. Similarly, high value of smoothness regularization enables smooth deformations with sub-optimal alignment. In practice, amongst other evaluation metrics, the Dice score and $|J_\phi|$ are used to access the result of the registration. In our method, we focus on these two metrics.

\textbf{Dice score:} 
Acquiring accurate human annotated anatomical segmentations is a tedious task, hence the annotations available are leveraged during training. For a well aligned image, regions in \emph{f} and \emph{m $\circ$ $\phi$} corresponding to the same anatomical region should overlap well. The Dice score quantifies the overlap between the structures. Let  \emph{s(f)$_i$} and  \emph{s(m)$_i$} represent labels of volumes \emph{f} and \emph{m} of structure \emph{i} respectively, we define the target Dice score as; 


\begin{equation}
\label{eqn:Dice}
\text{Dice}(\emph{s(f)$_i$},  \emph{s(m)$_i$}\circ \phi)  = 2 * \frac{| \emph{s(f)$_i$}  \cap ( \emph{s(m)$_i$} \circ \phi)|}{| \emph{s(f)$_i$}|  + | \emph{s(m)$_i$}\circ \phi|} =  \hat{y}_i
\end{equation}
 Where a score of 1 indicates a perfect match between anatomical labels, and 0 means there is no overlap. Thus, we can mathematically define $\mathcal{L}_{\text{overlap}}$ as MSE between the predicted Dice, $y_i$ and the target Dice, $\hat{y_i}$ given as;
\begin{equation}
\label{eqn:overlap}
\mathcal{L}_{\text{overlap}} = \frac{1}{|N|} \sum_{i \in N}(y_i - \hat{y}_i)^2
\end{equation}

Since we also aim to evaluate the effect of hyperparameter on each label, we predict 35 different Dice scores, each representing the degree of overlap for an anatomical region in the brain.

\textbf{Number of folded voxels:} We obtain the target number of folded voxels from the registration by computing the Jacobian determinant of the deformation field $|J_{\phi}|$ for each input, this gives us an understanding of its diffeomorphic properties. Where negative values of $|J_{\phi}|$ indicates non-plausible mapping, values less than 1 means contraction around that local region and values greater than 1 denotes expansion within that region. In practice, the displacement field is represented in an n + 1D image, meaning for each voxel p within the spatial domain, $\Omega$, the displacement u(p) represents a shift that aligns \emph{f}(p) and [\emph{m $\circ$ $\phi$}](p) to corresponding anatomical regions. The target number of folded voxels, $x_i$, is derived by computing the sum where $|J_{\phi}| < $ 0  acorss the entire displacement field. Hence, $\mathcal{L}_{\text{nfv}}$ in Equation \ref{eqn:lg} can be formally defined as;

\begin{equation}
\label{eqn:nfv}
    \mathcal{L}_{\text{nfv}} = \frac{1}{|N|} \sum_{i \in N}(x_i - \hat{x}_i)^2
\end{equation}

Where $x_i$ and $\hat{x_i}$ denote the predicted and target nfv respectively. Throughout our experiments, we express nfv as a percentage, i.e., \%nfv.

\section{Results}

We evaluate our method on 3D brain MR scans. To demonstrate our contributions, we train and evaluate HyperPredict models using two pre-trained convolutional encoders and two registration models as previously mentioned. Subsequent sections provide a comprehensive description of the experimental setup and the experiments conducted.

\noindent\textbf{Dataset and Pre-processing:} 
We demonstrate our method on brain MRI registration task. We use 414 T1-weighted brain scans from the OASIS dataset \citep{brains2020open, marcus2007open}. The dataset contains cross-sectional collection of subjects (men and women) aged from 18 to 96. The pre-processed version of OASIS dataset is obtained from \citep{hoopes2021hypermorph}. They performed standard pre-processing steps using FreeSurfer, \citep{fischl2012freesurfer} including; affine spatial normalization, skull stripping, sub-cortical structures segmentation, and finally, cropping the resulting images to 160 x 192 x 224.  The dataset includes sub-cortical segmentation maps of 35 different anatomical structures for each volume, generated by FreeSurfer. We consider Freesurfer a silver-standard method for generating automatic brain segmentations \citep{puonti2016fast}. We used the resulting segmentation maps during training to compute the target segmentation overlap between the image pair. We divide this dataset into train, validation and test sets of sizes 254, 80, and 80 respectively.

\noindent\textbf{Implementation:} 
To improve the computational efficiency of our experimental process, we precompute registrations using both cLapIRN and Niftyreg between all images pairs using randomly selected regularization weights, sampled from a log-Normal distribution.
We save the resulting metrics of these registrations in a csv file, which is used to train the MLP network. 

The parameterization of $g_{\theta}$ is based on three linear layers, each followed by a LeakyReLU \citep{maas2013rectifier} activation function except the final layer.
 Our proposed method is implemented in PyTorch, we adopt Adam optimizer, \citep{kingma2014adam} with a learning rate of $10^{-4}$. The training process follows the description in Section \hyperref[sec:HyperPredict]{4.1}.

 
We ran our model using both the SymNet and cLapIRN encoders. The results presented in Appendix \href{fig:summary_statistics}{C} demonstrates that SymNet achieved a lower Mean Absolute Error for both the Dice score and \%nfv. Hence, for all experiments detailed in this section, we show results for Symnet as an encoder. To experiment HyperPredict's ability to generalize across different registration methods, we show results for HyperPredict on cLapIRN and Niftyreg registration methods separately. We label HyperPredict trained with both methods as HyperPredict$_{\text {clap}}$ and HyperPredict$_{\text{nr}}$ respectively. We denote cLapIRN \emph{single} registration hyperparameter as $\lambda$. Niftyreg offers various regularization options, we use spacing, \emph{sx}, bending energy, \emph{be}, and linear elasticity, \emph{le} in our experiments.

\subsection{Experiment 1: Accuracy of HyperPredict}
This experiment aims to validate our proposed method by assessing if, and how accurately a HyperPredict model is able to predict the desired evaluation metrics. We check to see if through the registration algorithms, HyperPredict learns to capture the effect of a wide range of hyperparameter values simultaneously. We present a validation plot that shows our ability to predict over both metrics for selected values of hyperparameters on the test dataset. Additionally, we present Bland-Altman plots to evaluate the agreement between predicted values from HyperPredict and the resulting registration metrics.
\begin{figure}[ht!]
    \label{fig:predicted_vs_target}
    \centering 
    \includegraphics[width =1.025\textwidth]{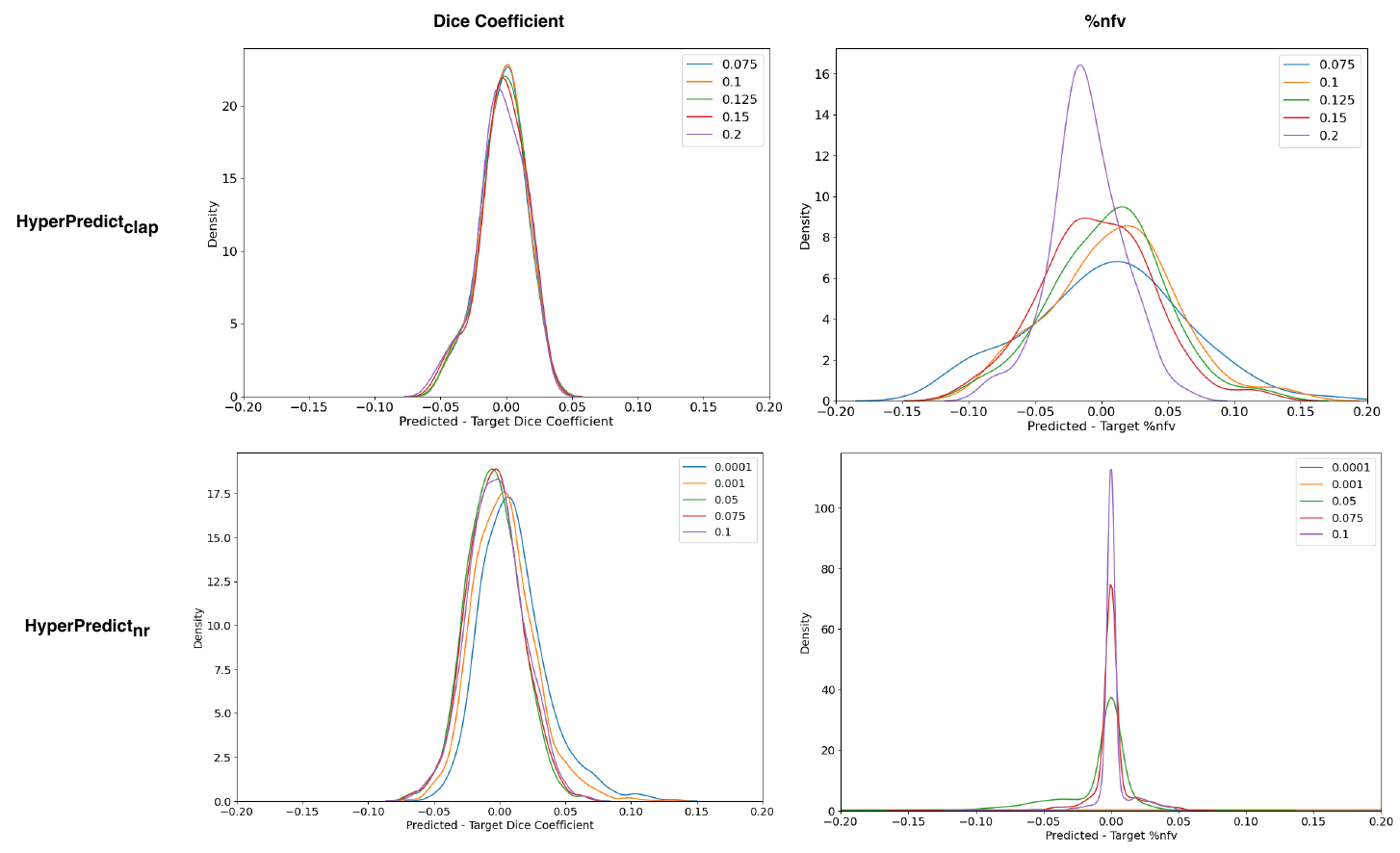}
    \caption{Error Distribution Plots on Both HyperPredict Models. \textbf{Left:} Difference in predicted and target dice coefficients for selected hyperparameter values. \textbf{Right:} Difference in predicted and target \%nfv for selected hyperparameter values. For visualization purpose, we display results for selected values.} 
\end{figure}

\begin{figure}[ht!]
    \label{fig:validating_hyperpredict}
    \centering 
    \includegraphics[width =1.025\textwidth]{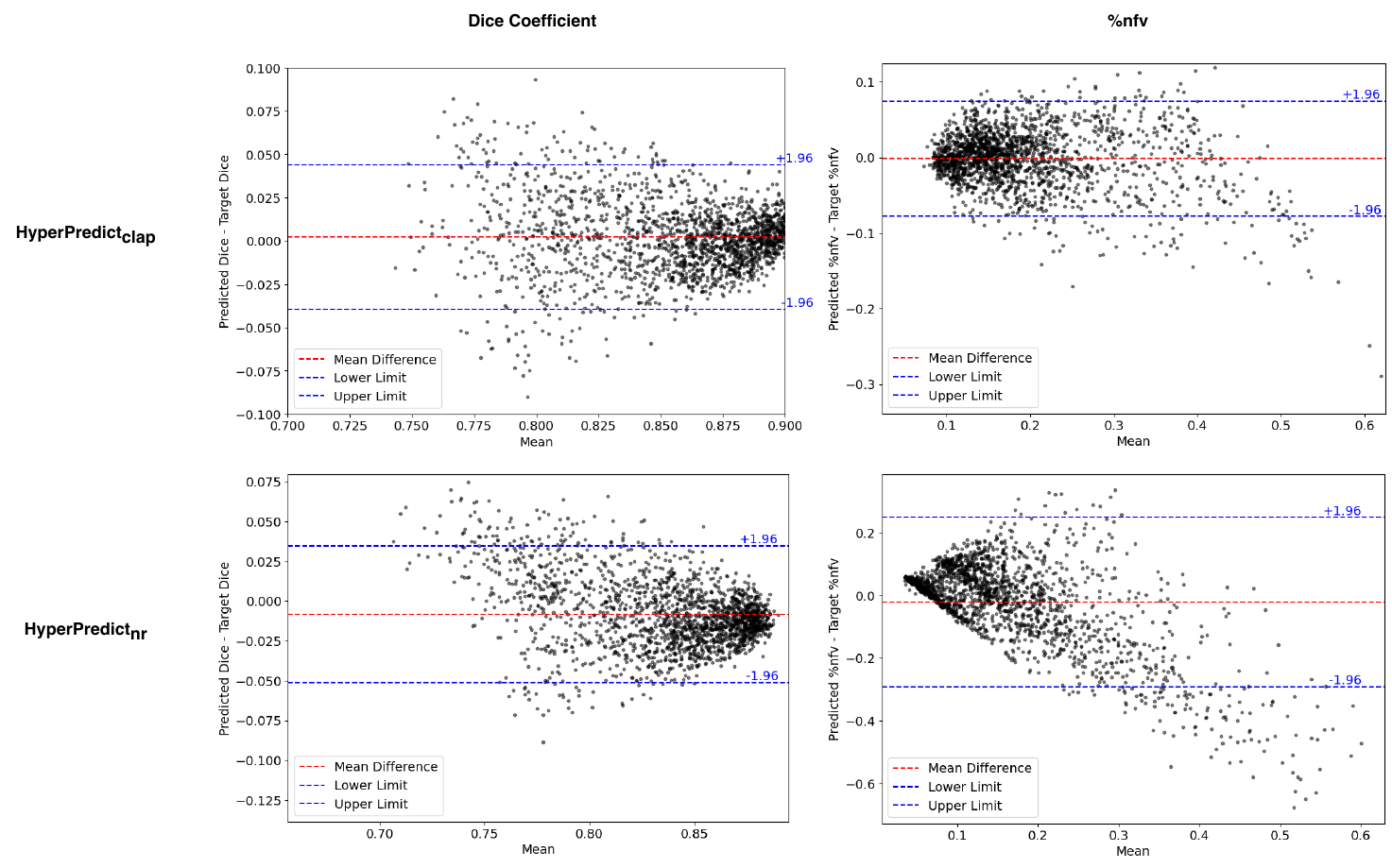}
    \caption{Validating HyperPredict. \textbf{Left:} Bland-Altman plots showing the agreement between the predicted vs target Dice scores across the entire population for both models (each point represents the average Dice for a particular image pair). \textbf{Right:} Bland-Altman plots showing the agreement between the predicted vs target \%nfv for the entire population. } 
\end{figure}

\noindent\textbf{Experiment Setup and Results:} First, we separately train two HyperPredict models, one for each registration algorithm (cLapIRN and Niftyreg) using the steps outlined in \hyperref[sec:method]{method} section. To measure the models predictive performance, we analyze the residual plots between the predicted and target metrics. To do this, we sample arbitrary discrete values of hyperparameters; $\lambda \in$ [0.05, 0.075, 0.1, 0.125, 0.150, 0.2, 0.5,  1], and \emph{be} $\in$ [0.0001, 0.001, 0.0075, 0.05, 0.075, 0.1, 0.125], we observed more variance with smaller values of bending energy. We use the sampled hyperparameter values to run image pairs in the test set on HyperPredict$_\text{{clap}}$ and HyperPredict$_\text{{nr}}$ respectively. Similarly, for each test pair and hyperparameter value, we derive the corresponding target metrics from both registration methods. Finally, for the range of hyperparameter values selected, we compute the difference between the predicted and target for both metrics on each HyperPredict model. We present the error distribution plots in Figure \hyperref[fig:predicted_vs_target]{2}. 

To assess the agreement between the predicted and target metrics, we use Bland-Altman plots. For HyperPredict$_\text{{clap}}$, we generate 200 log-linearly spaced values between -10 and 0, $\lambda \in [-10, 0]$. Similarly, for HyperPredict$_\text{{nr}}$, we sample 200 log-linearly spaced values for the bending energy, $be \in [-10, 0]$,  we use a fixed value of 5 for the spacing, $sx \in [5]$ (Segmentation overlap metrics indicate that the smallest spacing is better in Niftyreg, see Figure 9 (left) in Appendix A, but we compromise for computational efficiency), and a fixed value of 0.01 for linear elasticity, \emph{le}. For image pairs in the test dataset, with HyperPredict$_\text{{clap}}$, we run each of the 200 sampled hyperparameter value, $\lambda$ with the pair to obtain the predicted metrics. Next, we select the optimal $\lambda$ value per image pair and pass it to the cLapIRN registration to obtain the target metrics. The same procedure is done for HyperPredict$_\text{{nr}}$, using sampled \emph{be}, and fixed values of \emph{le} and \emph{sx} as parameters.

In Figure \hyperref[fig:validating_hyperpredict]{3} we present the pairwise differences between the predicted and target metrics plotted against their average, for both HyperPredict$_\text{{clap}}$ (top) and HyperPredict$_\text{{nr}}$ (bottom). The plots reveal a close agreement between results from the predicted and target metrics for HyperPredict$_\text{{clap}}$, with narrow upper and lower boundaries, indicating that majority of the differences fall within an acceptable range. Furthermore, the figure demonstrates that our method is able to yield metrics similar to those obtained through cLapIRN registration for each image pair. We found an average difference in the predicted and target Dice scores to be 0.015$\pm$0.014 and 0.028$\pm$0.026 for the \%nfv. 

Using HyperPredict$_\text{{nr}}$, we observe reasonable agreement between the predicted and target dice scores (bottom left), however, our model exhibits some limitations in accurately predicting \%nfv (bottom right). The determined difference is 0.018$\pm$0.013 and 0.099$\pm$0.109 for Dice and \%nfv respectively.

\subsection{Experiment 2: HyperPredict vs Cross-Validation}

The standard approach to obtaining an optimal hyperparameter is by performing cross validation. Hence, this section aims to evaluate the effect of inferring instance specific hyperparameters using HyperPredict. We achieve this by comparing the results obtained from our method with standard cross-validation.\\

\noindent\textbf{Experiment Setup and Results:} In this section, we conduct three experiments. 

\textbf{Step 1:} In the first experiment, we compare the Dice scores obtained using our method on selected anatomical structures and \%nfv with that of cross-validation. For HyperPredict$_\text{{clap}}$ and cross-validation on cLapIRN, we leverage the low computational burden offered by HyperPredict to test multiple hyperparameter values on a single pair. The sampling method is similar to the previous experiment. At test time, we run the sampled values for each subject in the test set through the trained HyperPredict$_\text{{clap}}$ network. Using the criteria defined in Equation \ref{eqn:constraint}, we select the optimal hyperparameters across specific labels. Finally, we register the image pair using the derived optimal values to obtain the corresponding target Dice coefficient and \%nfv.

\begin{figure}[ht!]
    \label{fig:hyperpredict_vs_cross_validation}
    \centering 
    \includegraphics[width =0.8\textwidth]{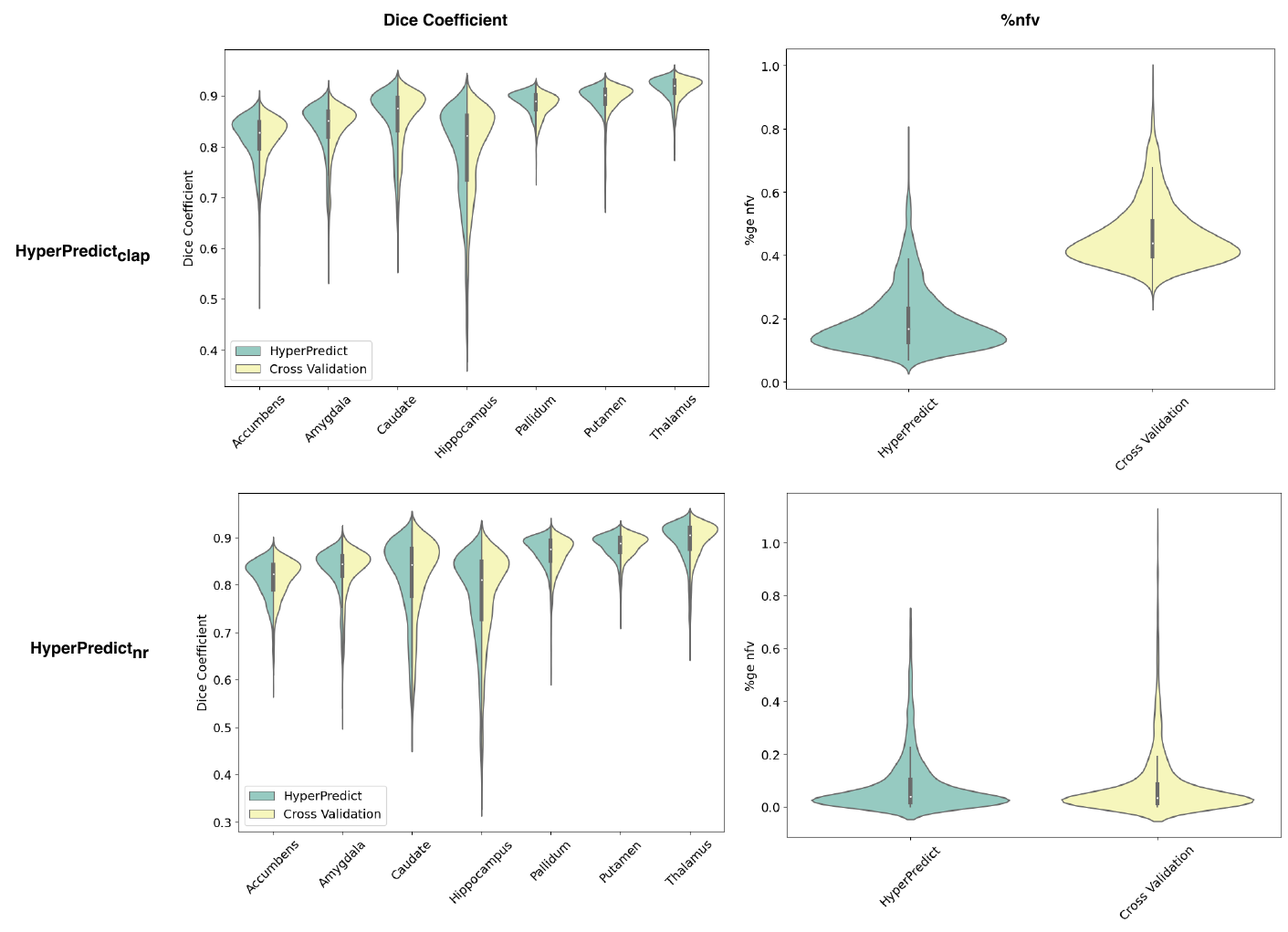}
    \caption{HyperPredict vs Cross-Validation. \textbf{Left:} Comparison of the Dice Coefficient on both methods for selected labels across the population. \textbf{Right:} Comparison of the \%nfv on both methods across the population. } 
\end{figure}

Since in practice it is not feasible to run 200 hyperparameter values per image pair on a registration algorithm in search of the optimal hyperparameter, we perform cross-validation using 8 discrete values of $\lambda$ selected arbitrarily, $\lambda$ $\in$ [0.05, 0.075, 0.1, 0.125, 0.150, 0.2, 0.5,  1], cross-validation on cLapIRN found optimal value, $\lambda^*$, to be 0.075. We use this for registration at test time across the \emph{entire} test dataset.

We repeat steps the above steps using HyperPredict$_\text{{nr}}$ and Niftyreg registration algorithm. For cross-validation, we maintain a fixed spacing value of 5, a fixed linear elasticity of 0.01, and discretely sample the bending energy, \emph{be} $\in$ [0.0001, 0.001, 0.0075, 0.05, 0.1, 0.2, 0.5, 1]. Cross-validation on Niftyreg found optimal value, \emph{be}, to be 0.0075.

We compare the results of HyperPredict and cross-validation for both experiments using the Dice coefficient scores and \%nfv. In Figure \hyperref[fig:hyperpredict_vs_cross_validation]{4} (top row), we present results of HyperPredict$_\text{{clap}}$ vs cross-validation on cLapIRN for both metrics. The bottom row of the figure depicts results obtained from Niftyreg. 

From the Figure, HyperPredict$_\text{{clap}}$ demonstrates comparable Dice scores with those obtained through cross-validation, with HyperPredict slightly outperforming cross-validation on a few anatomical structures. Interestingly, we find that our method of selecting optimal hyperparameter values leads to a reduction in folded voxels, indicating more desirable (plausible) registrations (top right). Similarly, the experiment with Niftyreg shows comparable dice scores between both methods, and the image pair with the highest \%nfv has a value less than that of cross-validation. 
\begin{figure}[ht!]
    \label{fig:best_case_worse_case}
    \centering 
    \includegraphics[width =0.8\textwidth]{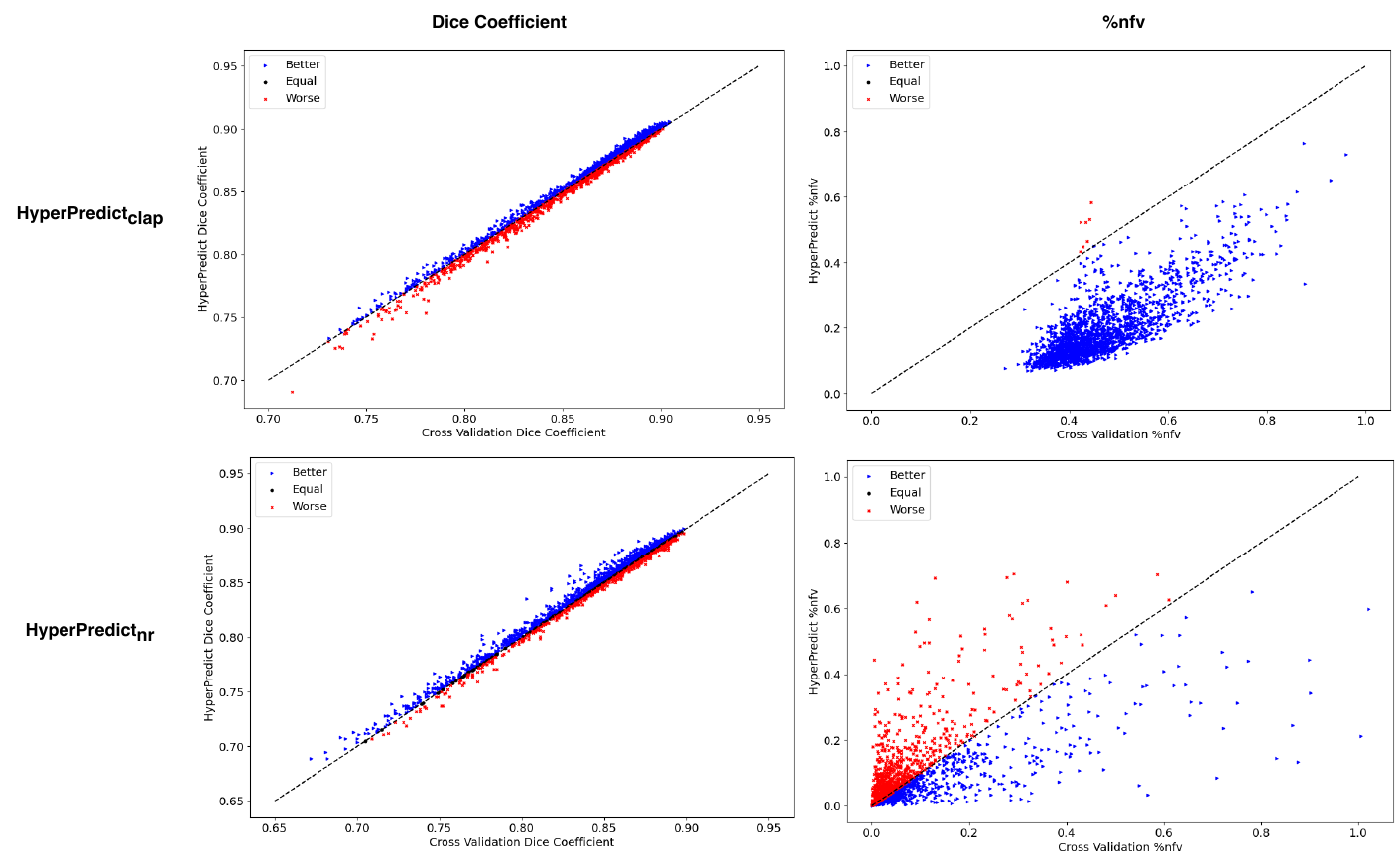}
    \caption{HyperPredict vs Cross-Validation. \textbf{Top:} Worse case analysis on HyperPredict$_\text{{clap}}$ \textbf{Bottom:} Worse case analysis on HyperPredict$_\text{{nr}}$. } 
\end{figure}

\textbf{Step 2:} For the second experiment, we perform a best and worst case analysis in Figure \hyperref[fig:best_case_worse_case]{5}, where we plot the value of the metric derived from HyperPredict against cross-validation for each image pair. We present results for HyperPredict$_\text{{clap}}$ and HyperPredict$_\text{{nr}}$. For all four plots depicted in the figure, the best cases are represented with a blue marker, indicating samples where HyperPredict out-performs cross-validation, red markers represent regions where cross-validation surpasses our proposed method, and equal cases are represented as black points that lie on the dotted slope.

\begin{figure}[ht!]
    \label{fig:clapirn_clapirn_exp2}
    \centering 
    \includegraphics[width =0.8\textwidth]{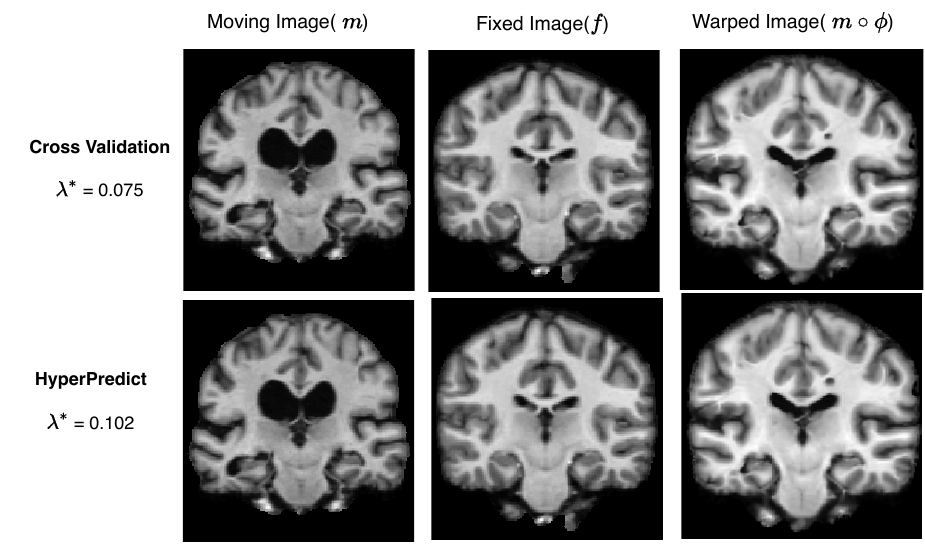}
    \caption{Cross Validation vs HyperPredict on cLapIRN registration: \textbf{Top:} Using optimal value of  $\lambda^*$ = 0.075 derived from cross validation to register an image pair. \textbf{Bottom:} Using optimal value of  $\lambda^*$ = 0.102 derived from HyperPredict$_\text{{clap}}$ to register the same image pair.}
\end{figure}

\textbf{Step 3:} In the final experiment comparing cross-validation with HyperPredict, we run the registration of a single image pair using the optimal $\lambda^*$ = 0.075 derived from cross-validation on cLapIRN. Similarly, we perform the registration on the same pair using the optimal $\lambda^*$ = 0.102 obtained from HyperPredict$_\text{{clap}}$ for that specific pair. The outcomes of these two approaches are illustrated in Figure \hyperref[fig:clapirn_clapirn_exp2]{6}. We record the dice score prior to registration as 0.425. The dice scores of both methods after registration are 0.755 and 0.751 respectively, and their corresponding folded voxels as 0.73\% and 0.48\%.

\subsection{Experiment 3: Analyzing the Distribution of Hyperparameters}
The objective of this experiment is to assess the distribution of regularization values for the two variants of HyperPredict: HyperPredict$_\text{{clap}}$ and HyperPredict$_\text{{nr}}$, and also identifying the best values to utilize for registration. We present the results obtained when optimizing for two instance-specific cases; across structures and across subjects. For visualization purposes, we present results of selected sub-cortical anatomical regions (Thalamus, Caudate, Putamen, Pallidum, Hippocampus, Amygdala, Accumbens).\\

\noindent\textbf{Experiment Setup and Results:} Maintaining the defined criteria and experimental setup in previous experiments, this section is subdivided into two distinct experiment.
\begin{figure}[ht!]
    \label{fig:kde_distribution_of_lamdas}
    \centering 
    \includegraphics[width =0.8\textwidth]{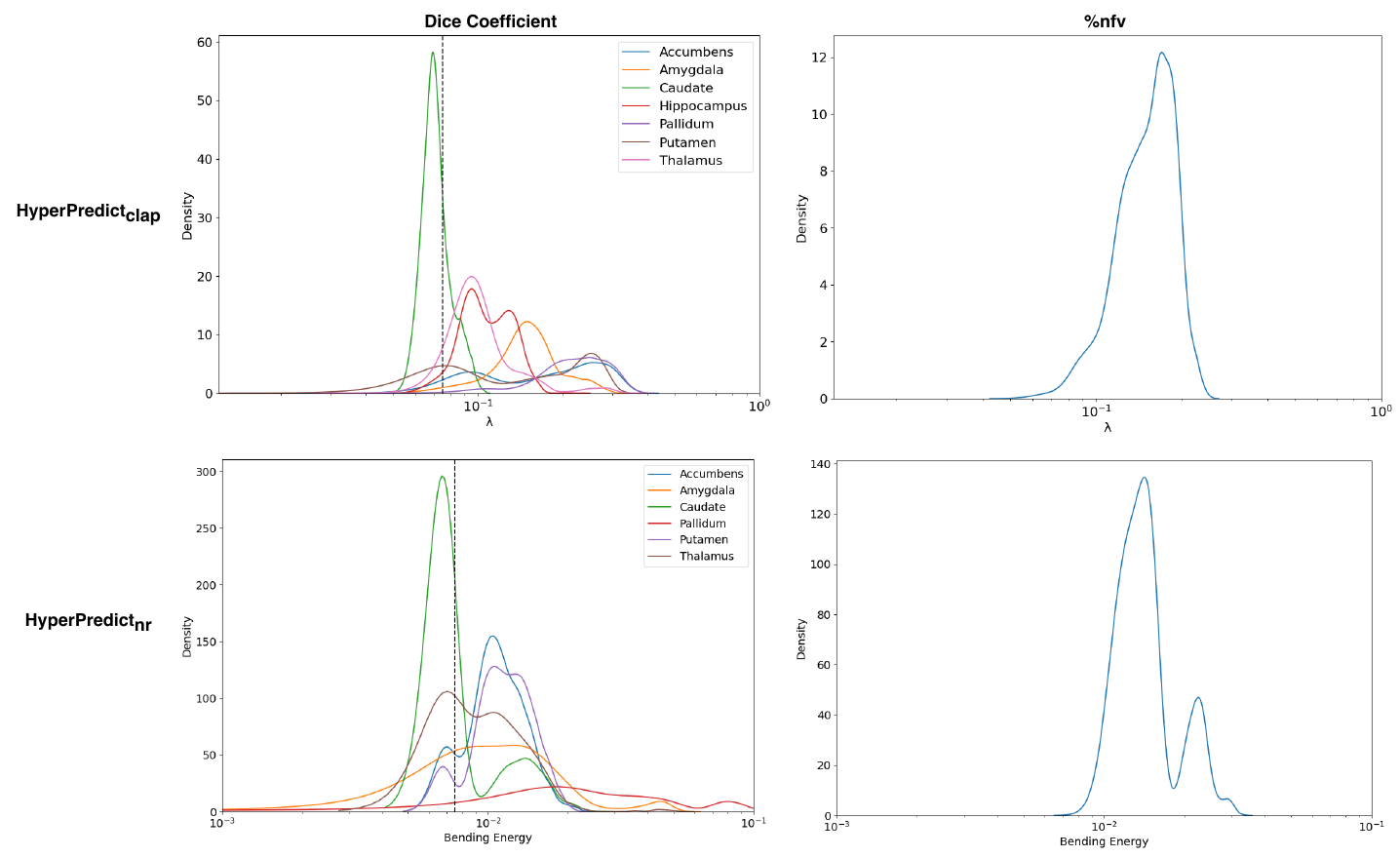}
    \caption{Distribution of Optimal Regularization Weights. \textbf{Left:} across labels \textbf{Right:} across subjects. For HyperPredict$_\text{{clap}}$(top row) and HyperPredict$_\text{{nr}}$(bottom row). The black-dashed lines in the plot represents the optimal value obtained through cross-validation method on both registration. } 
\end{figure}

\textbf{Step 1:} The first experiment seeks to validate our hypothesis that optimal values differ on a subject and structure basis. We infer the results of HyperPredict$_\text{{clap}}$ and HyperPredict$_\text{{nr}}$ using the test dataset and sampled values of $\lambda$ and \emph{be} respectively. From the result, we obtain the distribution of optimal hyperparameters across subjects (right) and selected labels (left) presented in Figure \hyperref[fig:kde_distribution_of_lamdas]{7}.

\begin{figure}[ht!]
    \label{fig:clapirn_clapirn_exp3}
    \centering 
    \includegraphics[width =1.0\textwidth, scale = 0.5]{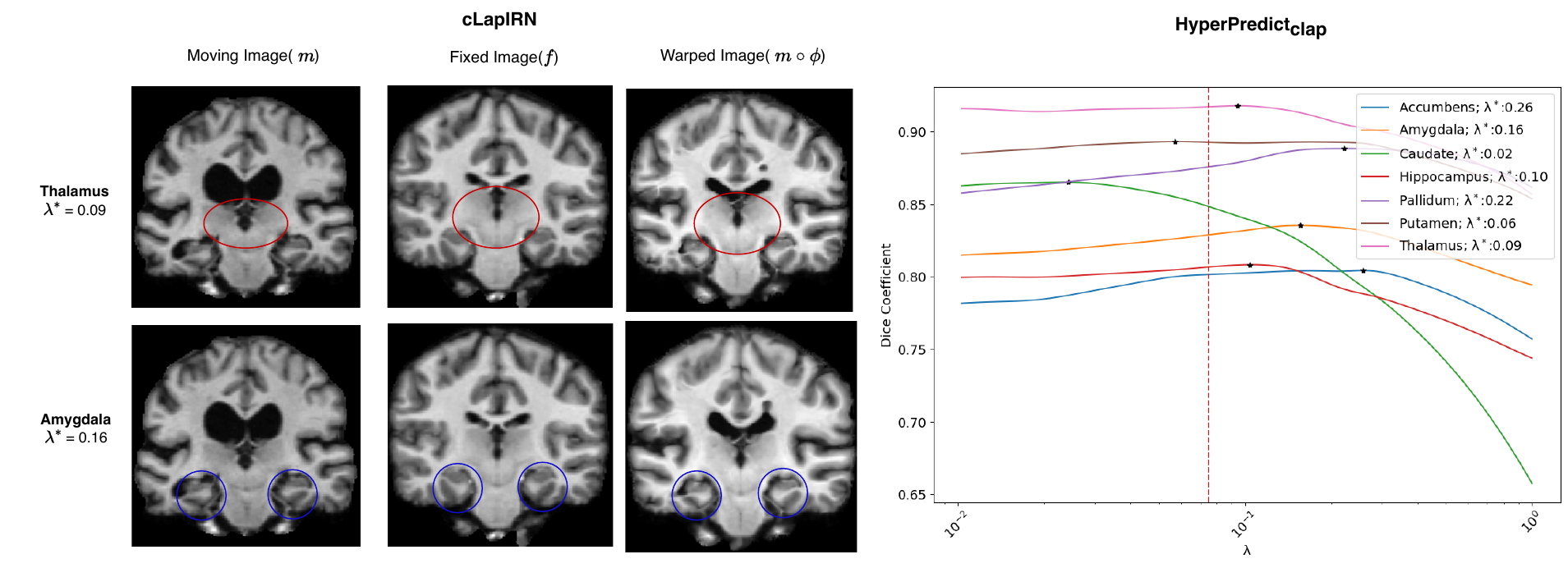}
    \caption{Varying regularization weights per anatomical structure. \textbf{Right:} Average Dice scores plotted against sampled regularization weights for selected anatomical regions. \textbf{Left:} Registration of an image pair using optimal parameters, with Thalamus and Amygdala as intended regions of interest.}
\end{figure}

\textbf{Step 2:} In the second experiment we identify label specific optimal regularization weights. For an image pair, HyperPredict derives 35 different Dice scores, each for an anatomical region. We leverage on this to observe the effect of utilizing instance-specific optimal values on corresponding regions. With the test results obtained from HyperPredict$_\text{{clap}}$, we compute the optimal Dice scores across each label. Figure \hyperref[fig:clapirn_clapirn_exp3]{8} (right) illustrates the varying optimal weights for selected structures. We visualize registration results (left) for two regions; Thalamus and Amygdala, using their optimal values, 0.09 and 0.16 respectively. In both registrations, we obtain an overall dice score of 0.75 and 0.73 across the entire image. However, when optimizing for Thalamus (using 0.09), the Dice score of Thalamus yields 0.89 and that of Amygdala was 0.80. Similarly, when optimizing for Amygdala (using 0.16), we obtained 0.78 and 0.80 for Thalamus and Amygdala respectively. Notice we obtain higher Dice scores for the specific region we optimize for, showing the significance of instance-specific regularization. The red-dashed line in the figure represents the optimal value obtained through cross-validation method, 0.075. We present result for HyperPredict$_\text{{nr}}$ in Appendix A, Figure 9 (right).

\subsection{Computational Efficiency at Test Time}
HyperPredict demonstrates notable computational efficiency at test time. A single model is capable of evaluating the effects of hundreds of hyperparameter values without imposing significant computational burden. In Table 1, we present a  summary that compares the test time of HyperPredict with other baseline registration methods.  It is important to note that while HyperPredict by itself is not a registration method, it acts as a tool that facilitates the selection of optimal hyperparameters. Therefore, the purpose of the comparison is to highlight that unlike HyperPredict, baseline methods are unable to infer the effect of hundreds (or thousands) of hyperparameters within a short period of time - which is a hindrance when trying to select the optimal hyperparameter. 

The runtime for methods under our proposed framework (utilizing Symnet encoder) is presented, with testing conducted using 8000 distinct hyperparameter values (HP) and includes the time it takes with and without registration (w/reg). HyperPredict with registration (w/reg) is the time it takes to run all 8000 values and using the optimal for registrating the image pair (either with cLapIRN or Niftyreg method), while HyperPredict without registration (w/o reg) is the time taken to test HyperPredict with 8000 values. We also present results of HyperMorph \citep{hoopes2021hypermorph}, highlighted in Section 3 as an additional baseline method, we asterisks it to show that the result is adopted from the paper.

\begin{table}[ht]
\centering
\caption{Comparison between HyperPredict and other registration methods at test time. Results are presented in seconds. }
\label{tab:test_time}
\begin{tabularx}{\textwidth}{X|X|X|X}
\hline
\multirow{2}{*}{\textbf{Method}} & \multirow{2}{*}{\textbf{Test Time(1HP)}} & \multicolumn{2}{c}{\textbf{Test Time(8000 HP)}}\\
& & w/o reg & w/reg\\
\hline
HyperMorph\text{*} & 2.1 & - & 16800 \\
\hline
Niftyreg  & 15.0 & - & 120000 \\
\textbf{HyperPredict$_\text{nr}$} & - & \textbf{0.84} & \textbf{15.84}  \\
\hline
cLapIRN  & 0.10 & - & 800 \\
\textbf{HyperPredict$_\text{clap}$} & - & \textbf{0.84} & \textbf{0.94}  \\
\hline
\end{tabularx}
\end{table}

\section{Discussion }

Our experiments demonstrated for the first time that the effects of a regularization hyperparameters can be predicted for pairwise image registration. We employ two metrics as a means of evaluating the registration; the dice score (for each segmented region), and the number of folded voxels (nfv), however, our framework could be extended to incorporate additional registration quality metrics.\\

\noindent\textbf{Accuracy of HyperPredict:} In our experiments, we utilize two registration algorithms: cLapIRN and Niftyreg to derive target values for training HyperPredict. Analysis from the residual plots in Experiment 1 reveals that the difference between the predicted and true values are centered around zero. This suggests that on average, our predictive model is making unbiased predictions and does not consistently overestimate or underestimate the predicted metrics. Similarly, the Bland-Altman plots show a good correlation between HyperPredict and cLapIRN registration. With Niftyreg, we find comparable results between the predicted and target dice coefficients, however, our model demonstrates sight limitations in accurately predicting the \%nfv. We hypothesize that this effect may be as a result of the use of a more complex similarity function (normalised mutual information) and the algorithmic heuristics, that makes it less predictable than cLapIRN. Another contribution to the sub-optimal predictive performance on \%nfv may be the choice of encoder. Currently, we employ two off-the-shelf convolutional encoders. The experiment presented in Appendix  C shows that the choice of encoder influences how effective HyperPredict is in estimating the desired metrics, it may be that: (a) the encoded representation itself does not contain sufficient information for HyperPredict$_{\text{nr}}$ to infer the \%nfv and/or (b) the use of a fixed convolutional encoder limits its ability to capture certain representations of the input specific to the task. In future, it is worth investigating the restrictions associated with encoder choice. \\

\noindent\textbf{HyperPredict vs Cross-Validation:} Cross-validation is the principle existing method for hyperparameter selection. The experiment results in Figure \hyperref[fig:hyperpredict_vs_cross_validation]{4} comparing HyperPredict and cross-validation show comparable dice scores using HyperPredict$_\text{clap}$ and HyperPredict$_\text{nr}$, with both HyperPredict approaches deriving slightly better results on a number of anatomical regions (Pallidum and Amygdala). We also find that HyperPredict$_\text{clap}$ produces registration with substantially less voxel folding than cross-validation (Figure \hyperref[fig:hyperpredict_vs_cross_validation]{4} top-right). This implies that our approach is able to capture the variability between subjects and identify an improved instance-specific level of regularization. 

We perform an additional comparison between both methods by investigating the proportion of examples where HyperPredict over-performs, under-performs, or yields comparable results to cross validation, presented in Figure \hyperref[fig:best_case_worse_case]{5}, we term this as \say{better}, \say{worse} and \say{equal} cases respectively in the Figure. For both HyperPredict$_\text{clap}$ and HyperPredict$_\text{nr}$, the Dice coefficient are comparable with HyperPredict$_\text{nr}$ having slightly more cases where HyperPredict performs better. HyperPredict$_\text{clap}$ demonstrates a significant improvement over cross-validation in terms of \%nfv, exhibiting lower \%nfv values for nearly all subjects (top-right). On the other hand, when comparing HyperPredict$_\text{nr}$ with cross-validation, there are slightly more instances where cross-validation outperforms HyperPredict$_\text{nr}$ (bottom-right), the reason for this may be attributed to the limitation in the predictive ability of HyperPredict$_\text{nr}$. 

Finally, the registration of the same image pair using the regularization weight of 0.102 and 0.075 derived from HyperPredict$_\text{clap}$ and cross-validation shows (in Figure \hyperref[fig:clapirn_clapirn_exp2]{6}) that while we arrive at a comparable dice score of 0.751 and 0.755 respectively, our method performs better in terms of \%nfv, with corresponding values of 0.48\% and 0.73\% respectively. This shows that utilizing HyperPredict enables us to arrive at more plausible deformation fields.\\

\noindent\textbf{Distribution of Regularization Weights:} Our study finds that different regularization weights yield optimal results depending on the subject and specific anatomical regions of interest. Our proposed framework has the flexibility of selecting such task-specific regularization values at test time as shown in Figures \hyperref[fig:kde_distribution_of_lamdas]{7} and \hyperref[fig:clapirn_clapirn_exp3]{8} (right), that can then be used for registration. The effect of such instance specific values is observed in Experiment 3 (Figure \hyperref[fig:clapirn_clapirn_exp3]{8}, left). When we apply the optimal regularization weight of independent anatomical regions (Thalamus and Amygdala) during registration, we derive higher dice scores for that particular region of interest.

We conduct an additional supplementary experiment presented in Appendix A that learns the effect of multiple hyperparameters. We utilize the Niftyreg algorithm due to its ability to incorporate a diverse range of regularization parameters, specifically we learn the effect of two of them simultaneously; bending energy (\emph{be}), and linear elasticity (\emph{le}). We present a heat map in Appendix A, Figure \hyperref[fig:heatmap]{10} depicting the combined impact of both parameters on the quality metrics across subjects. We observe from the figure that lower values of bending energy and linear elasticity results in our method predicting higher dice scores with more irregularities in the deformation field and vice-versa.\\

\noindent\textbf{Computational Efficiency:}  Finally, we analyse the computational efficiency of HyperPredict at test time. Exploiting a single trained HyperPredict model at test time means that for any arbitrary image pair, we are able to infer the effect of a range of hyperparameter values from a continuous interval without the need of labelled data while limiting the computational burden. It is important to note that our method is an \emph{amortized inference approach} that incurs an initial computational cost at train time, but results in a computationally efficient prediction at test time. Additionally, without our method a typical approach of selecting an optimal value at test time would involve running multiple independent registrations with different values. This approach becomes impractical as it restricts the search space of values and incurs additional computational costs for each independent registration.

Comparing the run time of a registration model using a single hyperparameter on an image pair to HyperPredict (testing thousands of values on the same image pair) reveals a notable difference in computational efficiency at test time. Table 1 shows that HyperPredict takes less than 0.84 seconds to predict the registration metrics for a single image pair with \emph{8000 different} hyperparameter values. This is approximately 8 times less the time it takes a cLapIRN registration model to derive the results from a \emph{single} hyperparameter value. The first column on the table presents the run time using a single hyperparameter value on various registration methods, the aim of HyperPredict is to test multiple values, hence we don't show results for our method in this column. In the second column, we show the run time with and without registration for 8000 distinct hyperparameter values. With HyperPredict, we can efficiently test and select the best value from all 8000 options. On the other hand, baseline registration methods necessitate more computational resources since the registration using all 8000 values need to be executed and the optimal one chosen based on the registration outcomes. Evidence of this is shown in the last column (w/reg), where it takes approximately 16800 seconds for HyperMorph to run 8000 distinct values on a single image pair, Niftyreg takes around 120000s (compared to HyperPredict$_\text{nr}$ that takes 0.84 seconds to run all 8000 values and an additional 15s to run registration on the optimal value). A similar pattern can be observed between cLapIRN and HyperPredict$_\text{clap}$, with runtime of 800 seconds and 0.94 seconds, respectively. HyperPredict enables a fast and more efficient method of hyperparameter selection at test time based on a flexible criteria, presenting a substantial advantage over baseline parameter selection methods.

\section{Limitations and Future Work}

The findings from HyperPredict demonstrate the efficacy of our approach in aiding the selection of hyperparameters that align with the unique attributes of the input. One limitation of our approach is the computational cost during training, as it involves running multiple registrations to optimize the MLP. Although, we perform further experiment presented in Appendix E that illustrates that HyperPredict can learn from 25\% of the available data while achieving satisfactory result. This significantly decreases the overall registrations that need to be run when utilizing our approach. To circumvent this computational cost at train time and for reproducibility, we ran multiple registrations (for both cLapIRN and Niftyreg) across all image pairs with sampled hyperparameter values. We have made the csv files containing registration results publicly accessible in our GitHub repository.

Our findings indicate that regularization weights vary across different anatomical structures, it is also likely that narrower regions in the brain are more difficult to register, future works will look into learning the importance of different regions and the effect it has on registration. Additionally, we plan to explore other hyperparameters, such as cost functions, as well as incorporating diverse datasets like Alzheimer's and lung data to generalize HyperPredict. Furthermore, experiments in Appendix B and C depicts the influence of model architecture and encoded representations on metric prediction, we intend to explore alternative encoded representations of image pairs. This can include deriving an image representation from a network that predicts the overlap between pairs of images, fine-tuning existing models, or utilizing self-supervised representation learning, we can then analyze the effect any of the methods have on the predicted result. Finally, in computing the number of folded voxels, we considered the entire image - including regions outside the labels, another problem formulation could be to learn the diffeomorphic properties of the image for only regions in the image where anatomical structures exist.

We have made our code, registration results and guidelines available on GitHub and we invite the community to contribute by replicating our  experiments on different datasets, conducting further evaluations on new experiments, or expanding the applicability of HyperPredict to different use-cases.

\section{Conclusion}

The accuracy of non-linear registration algorithms heavily depends on the choice of hyperparameter values, which may vary based on various factors. Thus, selecting a hyperparameter depending on specific use-case is an essential component of image registration. In this study, we propose HyperPredict, an efficient method of learning the effect of hyperparameters on registration, one that eliminates the need of running multiple registrations at test time in search for an optimal hyperparameter. HyperPredict utilizes an MLP that takes an image pair and desired set of hyperparameter as input and in return, predicts the evaluation metric that corresponds to the input. Our experiments show that by training a single HyperPredict model, we capture the effect of a range of hyperparameter values on an image pair. With this ability, at test time HyperPredict is able to predict the registration results of an image pair with thousands of registration hyperparameter values, giving us the ability to select the optimal value for a specific use-case. In summary, we have presented a novel approach we believe is more efficient in aiding the selection of optimal hyperparameter in image registration. While the experiments presented in this paper is specific to the domain of medical imaging, the general idea of HyperPredict can be adopted in other applications.




%
\ethics{The work follows appropriate ethical standards in conducting research and writing the manuscript. We train our models with publicly available data.}

\coi{We declare we don't have conflicts of interest.}



\clearpage
\appendix
\label{appendix}
\section{Experiments on Niftyreg}
\label{appendix_a}
	\noindent
    \begin{figure}[ht!]
    \label{fig:symnet_niftyreg}
    \centering 
    \includegraphics[width =0.9\textwidth]{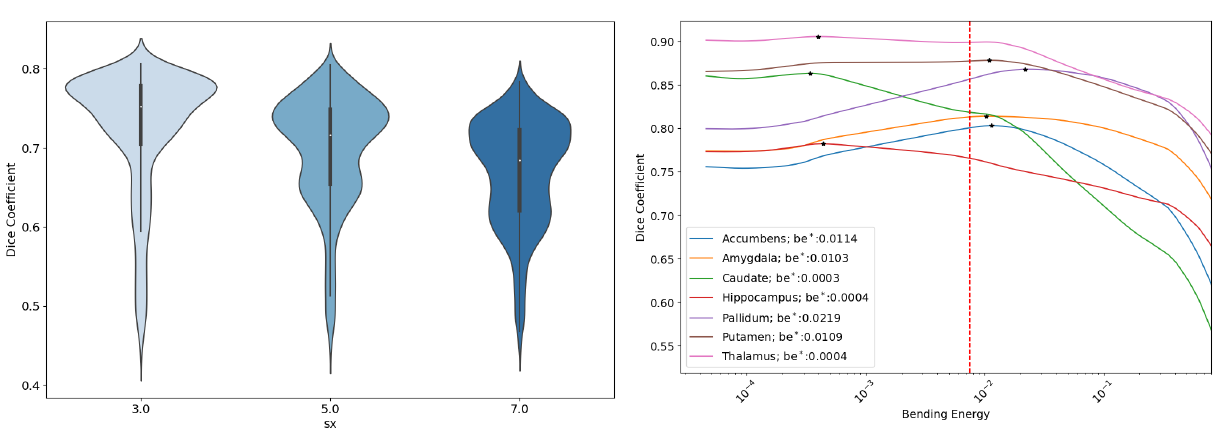}
    \caption{ \textbf{Left:} Niftyreg yields higher Dice coefficients with smaller spacing values. \textbf{Right:} Varying optimal bending energy across selected labels.}
\end{figure}

\begin{figure}[ht!]
    \label{fig:heatmap}
    \centering 
    \includegraphics[width =1\textwidth]{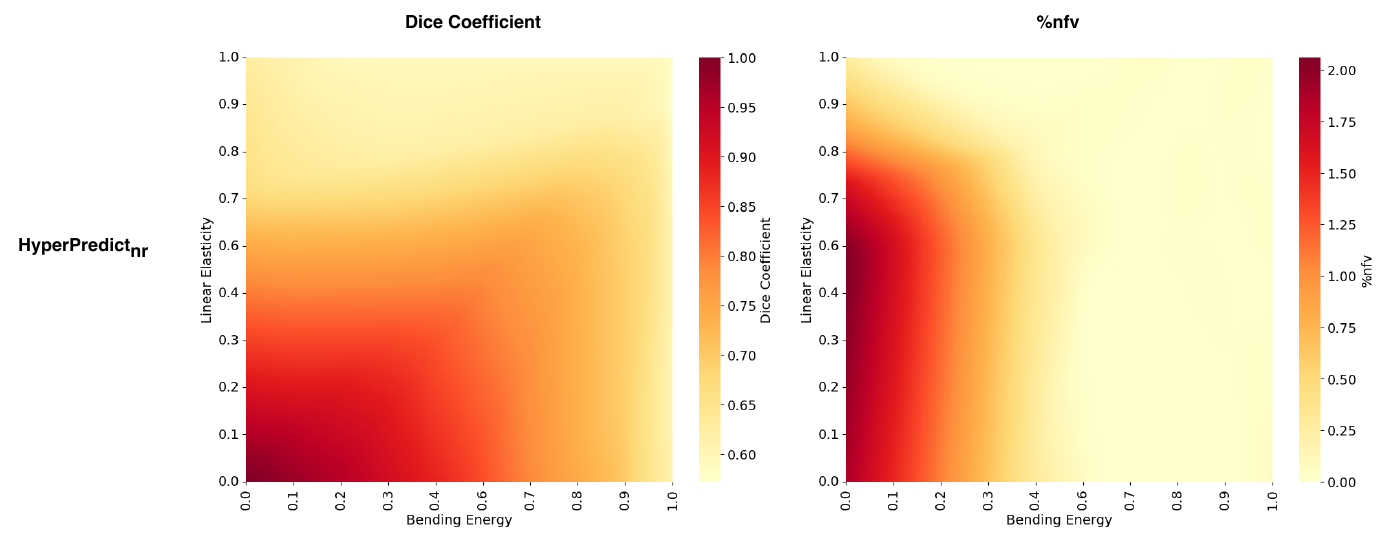}
    \caption{Heatmap depicting the combined effect of both hyperparameters; bending energy and linear elasticity on the dice score and \%nfv across subjects.}  
\end{figure}
Figure 10 is a supplementary experiment to investigate the combined effect of multiple regularization on each subject. To conduct this experiment, we train an additional HyperPredict$_\text{{nr}}$ model following the methodology outlined in Section 5. However, in this iteration the model specifically focuses on learning the influence of both the bending energy and linear elasticity. Both of which are sampled from a log-normal distribution and serve as input to the MLP at train time. At test time, for both \emph{be} and \emph{le}, we sample 200 log-linearly spaced values between -10 and 0 for each hyperparameter; this means each subject is tested with a combination of 40,000 values. Without a selection criteria, we present a heatmap above that shows the simultaneous effect of both bending energy and linear elasticity on the Dice coefficient and the folded voxels across subjects.

 \begin{figure}[ht!]
    \label{fig:non_constrained}
    \centering 
    \includegraphics[width =1\textwidth]{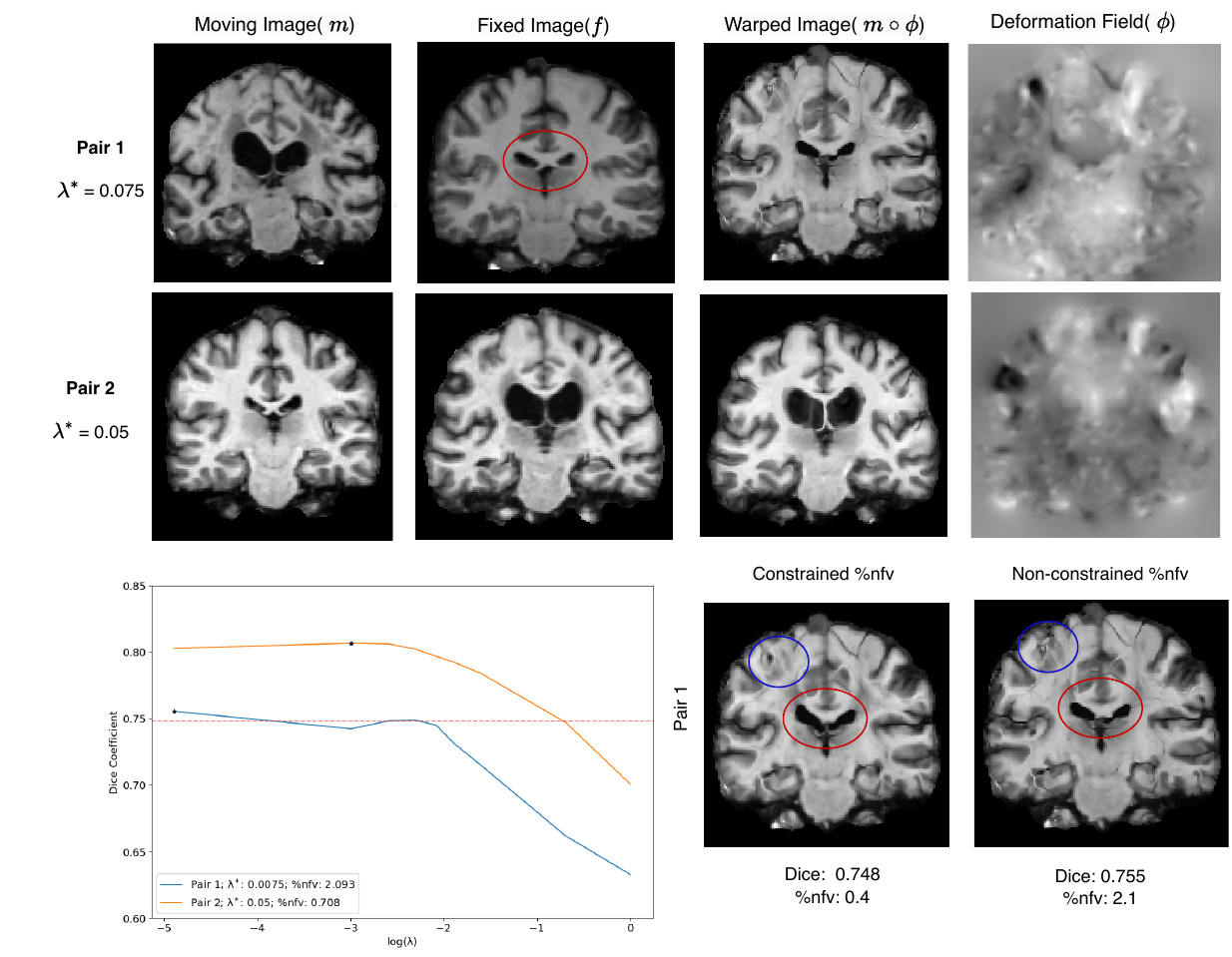}
    \caption{\textbf{Applying No Restriction on \%nfv}: Since the criteria used to define an optimal value is flexible after training, we change our criteria to consider no restrictions on the \%nfv in this experiment, and compare the results with the constrained version. \textbf{Bottom Left: }Dice scores plotted against values of $\lambda$, the red line is the optimal value inferred with restriction on \%nfv for this specific image pair (Pair 1). \textbf{Bottom Right: } Comparing warped images with and without restricted \%nfv for image Pair 1. The non-constrained warped image (brain image on the right) has a closer similarity to the fixed image compared to the constrained warped image (brain image on the left), this is made evident by the presence of a higher dice score (the red marker depicts visual similarities). However, we expect a less smooth warped image, as highlighted by the increased \%nfv from 0.4\% to 2.1\%, we depict this visually using blue markers.  }
\end{figure}


\clearpage
\section{HyperPredict Architecture}
	\noindent
\begin{figure}[ht!]
    \label{fig:exp5}
    \centering 
    \includegraphics[width =1\textwidth]{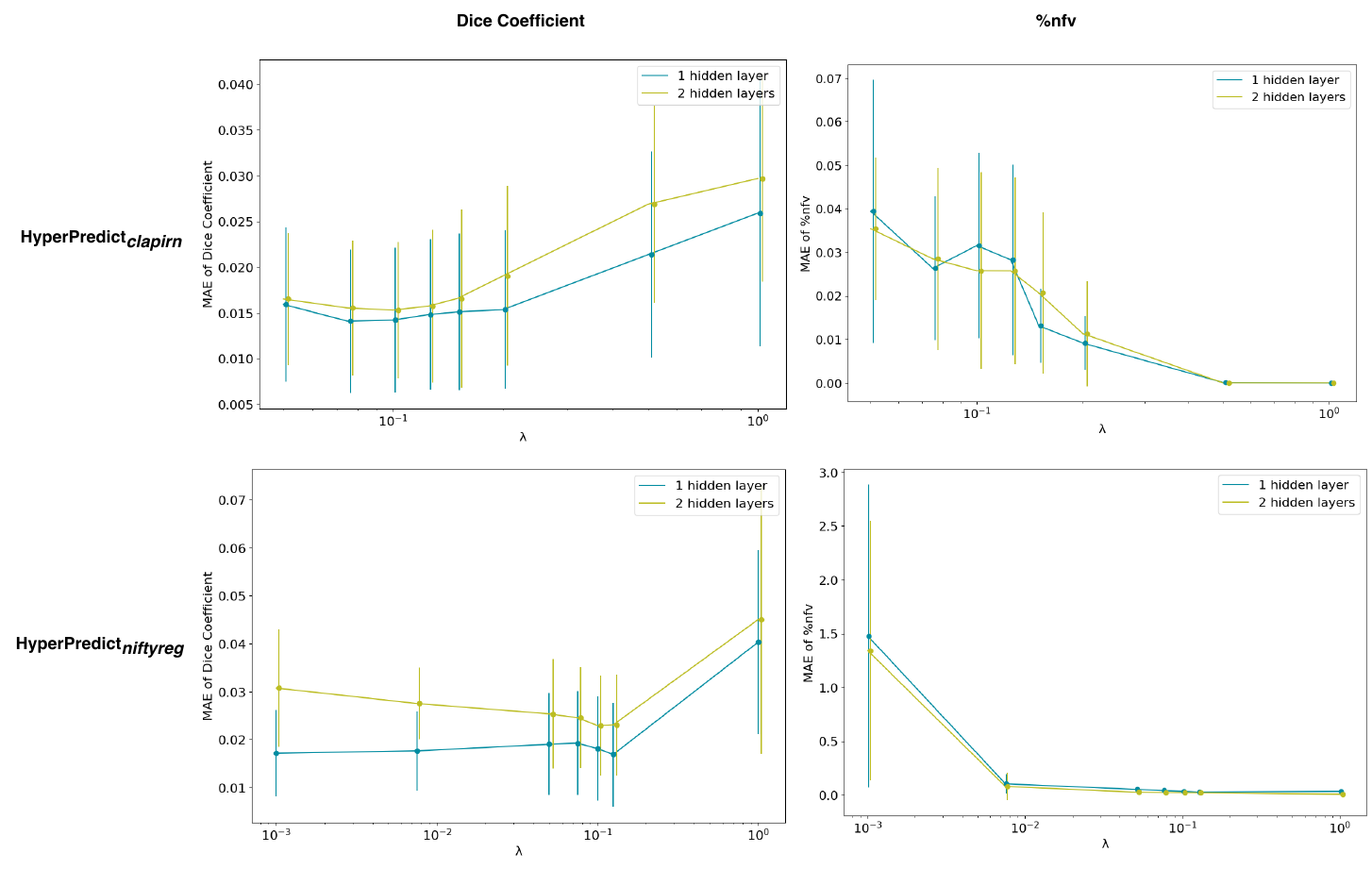}
    \caption{Effect of HyperPredict model on two different network architecture (using one and two hidden layers). \textbf{Top:} Results using HyperPredict$_\text{clap}$. \textbf{Bottom:} Results using HyperPredict$_\text{nr}$}.
\end{figure}

\clearpage
\section{Comparing Summary Statistics and Encoder type}
	\noindent
    \begin{figure}[ht!]
    \label{fig:summary_statistics}
    \centering 
    \includegraphics[width =1\textwidth]{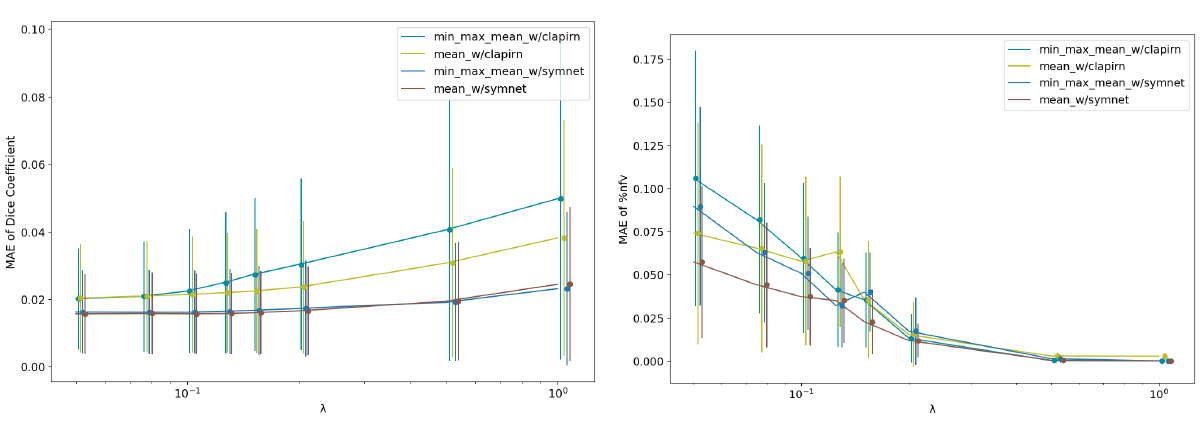}
    \caption{Comparing summary statistics and encoder types. Mean absolute error (MAE) of Dice scores (\textbf{left}) and \%nfv (\textbf{right}) using two different summary statistics (taking the \emph{mean} across the channels compared to concatenating the \emph{min}, \emph{max}, and \emph{mean} across the channels) and two different convolutional encoders (cLapIRN and Symnet).}
    \end{figure}

	\noindent
 \section{Combined Evaluation of Dice Coefficient and \%nfv}
  \begin{figure}[ht!]
    \label{fig:dice_vs_nfv}
    \centering 
    \includegraphics[width =1\textwidth]{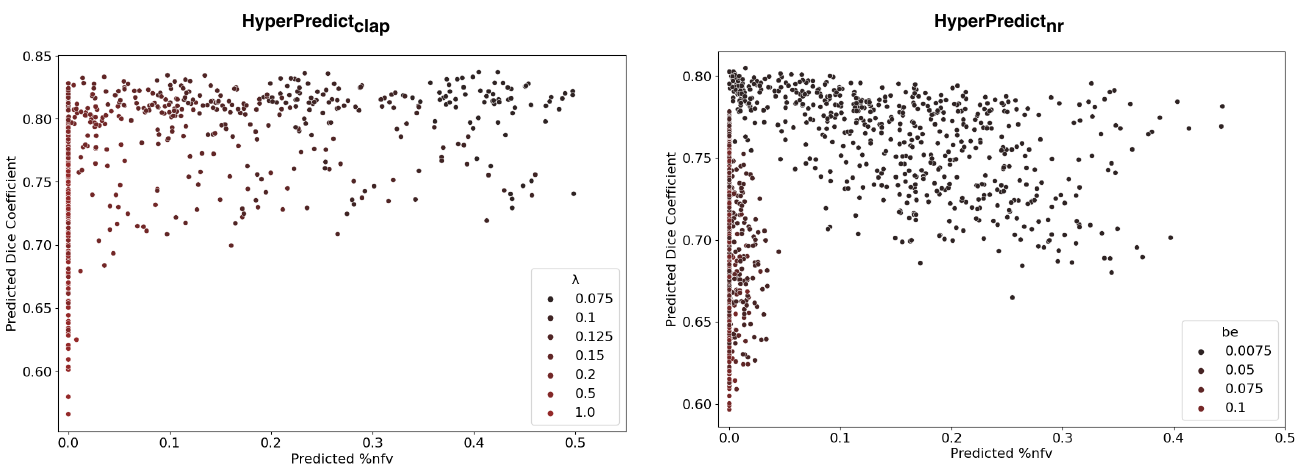}
    \caption{Distribution plot of Dice coefficient and \%nfv on selected values of $\lambda$ (left) and bending energy (right), for HyperPredict$_\text{clap}$ and  HyperPredict$_\text{nr}$ respectively. We show results of a sub-sample for visualization purpose. Both HyperPredict methods predict higher irregularities in the deformation field as the hyperparameter value gets smaller.}
\end{figure}

\clearpage

  \section{Data Size}
 \begin{figure}[ht!]
    \label{fig:datasize}
    \centering 
    \includegraphics[width =1\textwidth]{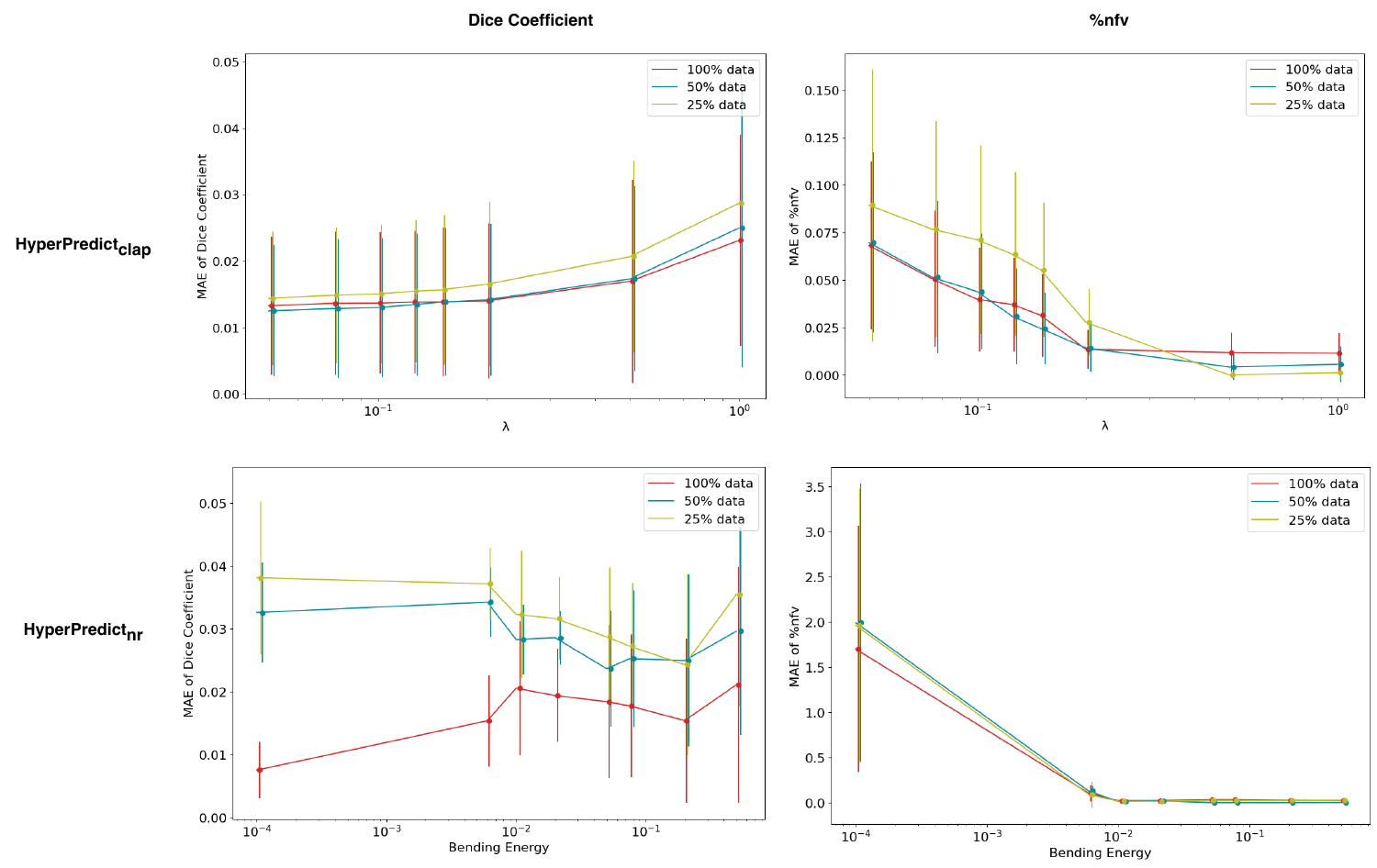}
    \caption{Training HyperPredict with different number of training samples (100\%, 50\% and 25\% of image pairs) \textbf{Top:} Mean absolute error of dice coefficient and \%nfv on HyperPredict$_\text{clap}$ \textbf{Bottom:} Mean absolute error of dice coefficient and \%nfv on HyperPredict$_\text{nr}$ }
\end{figure}
	\noindent

 \clearpage

  \section{Sensitivity Analysis}
 \begin{figure}[ht!]
    \label{fig:sensitivity_analysis}
    \centering 
    \includegraphics[width =1\textwidth]{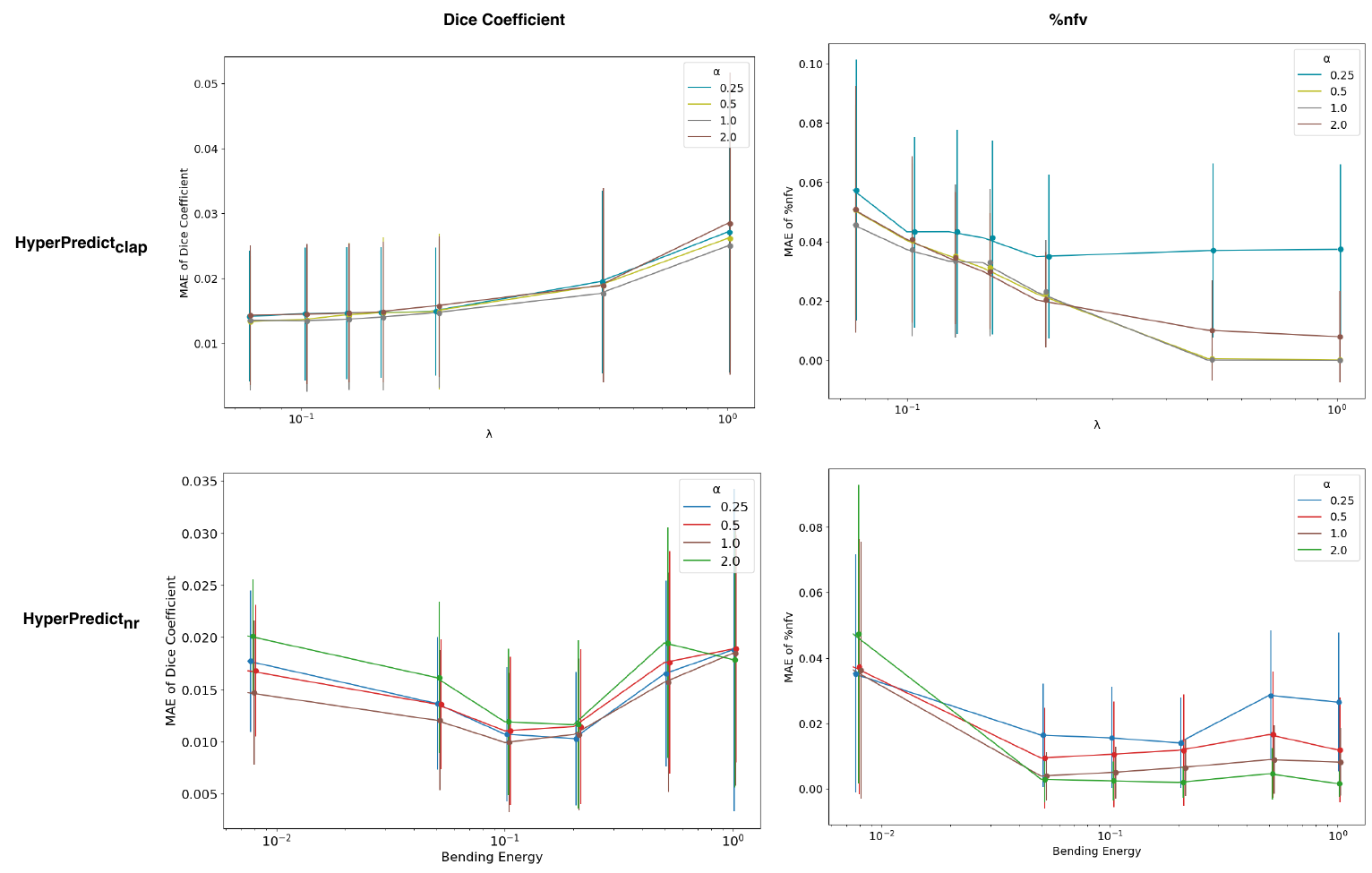}
    \caption{Sensitivity analysis result on $\alpha$ using varying values for training. We experimented with smaller values of $\alpha$, for visualization purposes, we present results for selected four. \textbf{Top:} MAE of dice coefficient and \%nfv on HyperPredict$_\text{clap}$ \textbf{Bottom:} MAE of dice coefficient and \%nfv on HyperPredict$_\text{nr}$}
\end{figure}
	\noindent

\end{document}